\documentclass[preprint,review,9pt]{elsarticle}

\usepackage{graphicx}
\graphicspath{{figures/}} 
\usepackage{amssymb}
\usepackage{enumitem}
\usepackage{multirow}
\usepackage{algorithm}
\usepackage{algpseudocode}
\usepackage{amsmath}
\usepackage{makecell}
\usepackage{xurl}
\usepackage{hyperref}

\usepackage{nameref}
\usepackage{booktabs}%
\usepackage[normalem]{ulem}

\newcommand{\Reals}{\mathbb{R}}

\newcommand{\PI}{\mathbb{N_+}}
\newcommand{\N}{\mathbb{N}}

\newcommand{\ROT}{\ensuremath{\mathtt{ROT}}}

\newcommand{\Sect}{Sec. }
\newcommand{\Table}{Table }
\newcommand{\degree}{^{\circ}}

\newcommand{\Fig}{Fig.}
\newcommand{\Eq}{Eq.}

\usepackage{soul}
\soulregister\cite7
\soulregister\ref7
\soulregister\pageref7

\journal{Pattern Recognition}

\begin{document}

\begin{frontmatter}

%% Title, authors and addresses

%% use the tnoteref command within \title for footnotes;
%% use the tnotetext command for theassociated footnote;
%% use the fnref command within \author or \affiliation for footnotes;
%% use the fntext command for theassociated footnote;
%% use the corref command within \author for corresponding author footnotes;
%% use the cortext command for theassociated footnote;
%% use the ead command for the email address,
%% and the form \ead[url] for the home page:
%% \title{Title\tnoteref{label1}}
%% \tnotetext[label1]{}
%% \author{Name\corref{cor1}\fnref{label2}}
%% \ead{email address}
%% \ead[url]{home page}
%% \fntext[label2]{}
%% \cortext[cor1]{}
%% \affiliation{organization={},
%%            addressline={}, 
%%            city={},
%%            postcode={}, 
%%            state={},
%%            country={}}
%% \fntext[label3]{}

\title{MS-TCRNet: Multi-Stage Temporal Convolutional Recurrent Networks for Action Segmentation Using Sensor-Augmented Kinematics
 }

%% use optional labels to link authors explicitly to addresses:
%% \author[label1,label2]{}
%% \affiliation[label1]{organization={},
%%             addressline={},
%%             city={},
%%             postcode={},
%%             state={},
%%             country={}}
%%
%% \affiliation[label2]{organization={},
%%             addressline={},
%%             city={},
%%             postcode={},
%%             state={},
%%             country={}}

\author[inst1]{Adam Goldbraikh\corref{cor1}}
\ead{goldb.adam@gmail.com}

\cortext[cor1]{Corresponding author}
\tnotetext[]{© 2024. This manuscript version is made available under the \href{https://creativecommons.org/licenses/by-nc-nd/4.0/}{CC-BY-NC-ND 4.0 license}}

\affiliation[inst1]{organization={Applied Mathematics Department at the Technion – Israel Institute of Technology},%Department and Organization
            city={Haifa},
            postcode={3200003}, 
            country={Israel}
            }

\author[inst2]{Omer Shubi}
\ead{shubi@campus.technion.ac.il}

\affiliation[inst2]{organization={Faculty of Data and Decision Sciences at the Technion – Israel Institute of Technology},%Department and Organization
            city={Haifa},
            postcode={3200003}, 
            country={Israel}}

\author[inst2]{Or Rubin}
\ead{orrubin@campus.technion.ac.il}

\author[inst3]{Carla M~Pugh}
\ead{cpugh@stanford.edu}

\author[inst2]{Shlomi Laufer}
\ead{laufer@technion.ac.il}

\affiliation[inst3]{organization={Stanford University, School of Medicine Stanford},%Department and Organization
            city={Stanford},
            state={CA},
            country={USA}}

\begin{abstract}

Action segmentation is a challenging task in high-level process analysis, typically performed on video or kinematic data obtained from various sensors. This work presents two contributions related to action segmentation on kinematic data. Firstly, we introduce two versions of Multi-Stage Temporal Convolutional Recurrent Networks (MS-TCRNet), specifically designed for kinematic data. The architectures consist of a prediction generator with intra-stage regularization and Bidirectional LSTM or GRU-based refinement stages. Secondly, we propose two new data augmentation techniques, World Frame Rotation and Hand Inversion, which utilize the strong geometric structure of kinematic data to improve algorithm performance and robustness. We evaluate our models on three datasets of surgical suturing tasks: the Variable Tissue Simulation (VTS) Dataset and the newly introduced Bowel Repair Simulation (BRS) Dataset, both of which are open surgery simulation datasets collected by us, as well as the JHU-ISI Gesture and Skill Assessment Working Set (JIGSAWS), a well-known benchmark in robotic surgery. Our methods achieved state-of-the-art performance.\\
\small
code:
\url{https://github.com/AdamGoldbraikh/MS-TCRNet}
\normalsize

\end{abstract}

%%Graphical abstract
% \begin{graphicalabstract}
% %\includegraphics{grabs}
% \end{graphicalabstract}

%%Research highlights


\end{frontmatter}

%% \linenumbers

\section{Introduction}\label{sec:introduction}
\subsection{Background}

The problem of action segmentation is one of the most challenging in high-level process analysis. Essentially, action segmentation involves labeling each timestamp in a temporally untrimmed sequence to create a segmented sequence.
Traditionally, action segmentation is applied to video data \cite{li2020ms}, kinematic data gathered from diverse sensors \cite{van2020multi}, or through multimodal approaches that integrate various data types \cite{van2022gesture}.  
Recently, it was demonstrated that action analysis can be conducted using a time series of 3D point cloud data  \cite{ben20243dinaction}.

Automatically segmenting long, untrimmed data sequences plays a crucial role in understanding human-to-human and human-to-robot interactions. By identifying actions, their initiation times, progress, environmental transformations, and future actions, these methods offer significant benefits. This understanding can impact various applications such as video security and surveillance systems, human-robot interaction, assistive systems, and automatic video summarization systems, enhancing the quality of everyday life.

In the context of surgical procedures, action segmentation is typically used as part of workflow analysis algorithms. These algorithms, implemented for minimally invasive surgeries (MIS) like laparoscopic, robotic-assisted (RAMIS) \cite{van2022gesture}, and open surgery \cite{goldbraikh2022using}, typically utilize video data, kinematic data, or both.

There are several advantages to using sensor kinematics over video data. First, in terms of runtime considerations, video-based action segmentation requires a computationally demanding stage of feature extraction from images. In contrast, using kinematic data eliminates this step, making the process more efficient. Second, from a privacy perspective, kinematic data does not involve capturing images of users, thereby avoiding potential privacy issues associated with video recordings.

 Action analysis based on kinematic data is used in surgical workflow analysis but is not limited to this domain; there is a growing interest in its application to industrial work analysis \cite{sopidis2022micro}, musician performance analysis \cite{fujisaki2022tool}, action recognition in sports \cite{hoelzemann2023hang}, and more. Several types of sensor data are used for this purpose. 

The first type includes sensors that report their pose relative to an external world frame, such as electromagnetic tracking sensors or optical tracking systems. These sensors provide precise kinematic data. However, they are restricted to very defined environments, limiting their use to controlled conditions.

Another popular kinematic sensing system is the inertial measurement unit (IMU), which includes an accelerometer, gyroscopes, and sometimes magnetometers \cite{ashry2020charm}. These sensors measure acceleration, angular velocity, and the Earth's magnetic field. By fusing this data, the sensor's orientation can be calculated. However, position and linear velocity estimation with IMUs suffer from drift over time, requiring complementary data for accuracy.

Data augmentation leverages existing data to create additional instances, enhancing algorithm generalization through increased data quantity. Typically, this process involves making minor modifications to the data. Traditional approaches focus on classical computer vision techniques such as cropping, zooming, flipping, and rotation. More recent advancements explore sophisticated methods, including style transfer and generative models \cite{ wang2024generative}.

Enhancing time series data, particularly when incorporating 3D spatial information, presents a significant challenge. Approaches for augmenting time series data encompass transformation-based techniques, decomposition methods, pattern mixing, and generative models \cite{iwana2021empirical}. In \cite{itzkovich2019using}  the use of geometric factors for augmenting kinematic data was first considered.

\subsection{Our Contribution}

This work aims to achieve two goals related to action segmentation tasks on kinematic data: the creation of new architectures and the development of data augmentations that are geometrically oriented. We demonstrate that our methods can be applied in two domains: the analysis of electromagnetic sensor motion data and robotic surgery kinematics records.

Firstly, we introduce two versions of Multi-Stage Temporal Convolutional
Recurrent Networks (MS-TCRNet): the bidirectional LSTM-based refinement, L-MS-TCRNet, and the bidirectional GRU-based refinement, G-MS-TCRNet. Both are specifically designed for kinematic data and demonstrate state-of-the-art results on benchmark datasets.
The main contributions of our new architectures are as follows:
\begin{itemize}
    \item A prediction generator with intra-stage regularization.
     \item  BiRNN-based refinement stages.
\end{itemize}

Similar to \cite{park2022maximization}, we adopted a divide-and-conquer approach to enhance both frame-wise accuracy and segmentation performance.
Intra-stage regularization is a technique that involves adding prediction heads after the internal dual dilated layers of the prediction generator. This improves the frame-wise performance of the network. In the BiRNN-based refinement stages, downsampling is applied to the input before it is fed into the BiRNN unit, and upsampling is applied to the output. As a result, the network's segmental performance is significantly improved and over-segmentation errors are greatly reduced.

To achieve the second goal, we propose two new data augmentation techniques for kinematic data, which take advantage of the data's strong geometric structure to improve the performance and robustness of algorithms. Specifically, we propose:

\begin{itemize}
    \item World Frame Rotation augmentation.
    \item Hand Inversion augmentation.
\end{itemize}
 The World Frame Rotation augmentation is inspired by image rotation. Based on three random Euler angles in a predetermined continuous range, we calculate an augmentation rotation matrix that multiplies the sensors' location coordinates and its original rotation matrix, which determines its real orientation. Since the same rotation matrix rotates all samples in the time series from all sensors, this is equivalent to rotating the world coordinate system.
 
Our Hand Inversion augmentation is inspired by a common horizontal flip of images, adapted to kinematic sensors.
Here, we calculate a plane that optimally separates the surgeon's right and left hand over time, reflect the data points across this plane, and exchanges each sensor's ID with the corresponding sensor from the other hand.

%%%%%%%%%%%%%%%%%%%%%%%%%%%%%%%%%%%%%%%%%%%%%%%%%%

\section{Related Work}
\label{sec:Related_Work}

\subsection{Action Segmentation}

Early methods for temporal action segmentation, like Bayesian non-parametric models, classified video sequences as visual words \cite{cheng2014temporal}. However, they struggled with long temporal contexts. Recurrent neural networks (RNNs) were proposed to capture long-term relationships in video sequences \cite{donahue2015long}. The latest methods use transformer models \cite{ke2024u} and Temporal Convolutional Neural networks (TCN) \cite{lea2016temporal}. MS-TCN++ \cite{li2020ms}, a multi-stage temporal convolutional network for action segmentation, uses video data input. It has a prediction generator and multiple refinement stages, each outputting a prediction. Input consists of feature vectors for each frame, typically extracted from a 3D CNN.
 In action segmentation, there is a trade-off between frame-wise and segmentation performance. To address this, \cite{park2022maximization} propose a TCN-based network that employs a divide-and-conquer method. This approach initially maximizes frame accuracy and subsequently reconstructs features to minimize over-segmentation.

\subsection{Kinematic Data and Analysis}
 Classical approaches, such as probabilistic graphical models, relied mainly on local transitions and thus missed relationships between long-range temporal events \cite{mavroudi2018end}.
 In \cite{forestier2018surgical}, continuous kinematic data was first discretized, both in the time domain and in the sensor values, into predefined bins and segments. After applying additional transformations, these segments were classified using cosine similarity.
 
Another approach that uses surgical kinematic signals as input was proposed in \cite{van2020multi}. Their method utilizes a multi-task RNN, predicting surgical gestures and surgical task progress in parallel.
Performing both maneuver and gesture recognition, \cite{dipietro2019segmenting} assesses a variety of Recurrent Neural Networks (RNNs), specifically simple RNNs, LSTMs, GRUs, and mixed-history RNNs.

Lea \textit{et~al.} \cite{lea2016temporal} introduced TCNs, which are designed to capture both long-range dependencies and detailed relations between timestamps more efficiently than RNNs. Goldbraikh \textit{et~al.} \cite{goldbraikh2022using} analyzed data from open surgery simulators using 6 degrees of freedom (DOF) motion sensors on surgeons' hands, the first to use kinematic data for recognizing gestures and tools. They established strong baselines for several models, with MS-TCN++ outperforming others in gesture recognition.

Other approaches utilize IMUs. In \cite{ashry2020charm} sensors in smartwatches were used to measure linear acceleration and angular velocities. After extracting features from the raw data, a \emph{Bidirectional Long short-term memory network} (BiLSTM) was used for classification. Similarly, \cite{huan2021human} used multiple on-body IMU sensors as input to a hybrid CNN-LSTM model.

\subsection{Temporal Data Augmentations}
Various techniques exist for augmenting general time-series data, from simple transformations to advanced methods like generative models and upsampling approaches \cite{semenoglou2023data}. In \cite{rashid2019times}, augmentations for IMU sensors on construction equipment include temporal modifications such as jitters, scaling, sensor rotations, and time warping. Regarding surgical gesture recognition and improving generalization between dry-lab and clinical-like data, \cite{itzkovich2019using} proposes rotating sensors individually in specific axes. Additionally, \cite{itzkovich2021generalization} suggests augmenting datasets by swapping data between right- and left-hand surgeons to address the imbalance in handedness representation.

\section{Datasets}
\label{section:datasets}

\begin{figure*}[ht]
 \centering
 \includegraphics[width=1\textwidth]{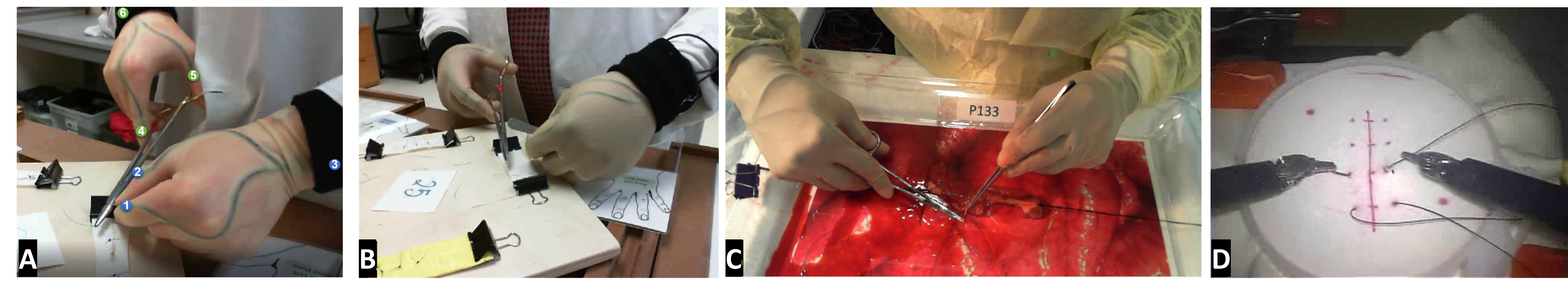}
 \caption{
 (A) Participants' hands with sensors, from the VTS dataset. In BRS the sensors are positioned in the same locations.\\
 The three datasets used to evaluate our algorithms and augmentations, (B) VTS - Variable Tissue Simulation Dataset, (C) the Bowel Repair Simulation Dataset, and (D) JHU-ISI Gesture and Skill Assessment Working Dataset.}
    \label{fig:datasets}
 \end{figure*}

We evaluated the proposed models on three different datasets of surgical suturing task: the Variable Tissue Simulation (VTS) Dataset \cite{goldbraikh2022video}, the Bowel Repair Simulation (BRS) Dataset \cite{basiev2022open}, and the JHU-ISI Gesture and Skill Assessment Working Set (JIGSAWS) \cite{gao2014jhu} (see \Fig \ref{fig:datasets}). VTS and BRS are open surgery simulation datasets collected by us, while JIGSAWS is a well-known benchmark that contains three elementary surgical training tasks:  suturing, knot tying, and needle passing, performed on the Da Vinci Surgical System simulator. The BRS dataset is presented here for the first time in the context of action segmentation by kinematic data.
Firstly, we describe the unique parts of the VTS (\Sect \ref{section:dataset_VTS}) and BRS (\Sect \ref{section:dataset_BTS}) datasets, then in \Sect \ref{subsec:acquisition} we describe the shared technical components. Finally, in \Sect \ref{subsec:JIGSAWS}, we describe the JIGSAWS dataset.

\subsection{Variable Tissue Simulation (VTS) Dataset}
\label{section:dataset_VTS}

  A simulated suturing task was performed using the variable tissue simulator \cite{goldbraikh2022video}. Two sections of material were stitched using three interrupted instrument-tied sutures. The materials were tissue paper and rubber balloons, with two repetitions for each. Eleven medical students, thirteen attending surgeons, and one resident completed 100 procedures, each lasting 2-6 minutes. The VTS dataset identified six suturing gestures: background, needle passing, suture pulling, instrumental tie, knot laying, and suture cutting.

\subsection{Bowel Repair Simulation (BRS) Dataset}
\label{section:dataset_BTS}

Data was collected using a simulator representing an operating room trauma patient with an abdominal gunshot wound. The dataset comprises 255 porcine enterotomy repair procedures performed by surgeons of various skill levels. We defined five surgical maneuvers: suture throws, instrument knot ties, hand knot ties, thread cuts, and background maneuver. Among participants, 52 performed a single-layer repair on the large hole, with only the suturing part annotated, resulting in 52 sequences in this dataset. Of the 52 participants, 45 were right-handed, three were left-handed, three were ambidextrous, and one had unknown handedness. Detailed dataset analysis is in Supplementary Materials \ref{sub_sec:BRS_dataset_analysis}.

\subsection{VTS and BRS Data Acquisition and Prepossessing}
\label{subsec:acquisition}

Motion data was captured using electromagnetic sensors (NDI, trackSTAR Model 180) attached to participants' index, thumb, and wrist under surgical gloves (see \Fig \ref{fig:datasets}- A). Additionally, video data was captured, with two synchronized cameras: one close-up on the simulation area and one overview, using MotionMonitor software (Innovative Sports Training, Inc.).

Labeling relied on the video data, defining activities by start and end times. Each sensor provided three spatial coordinates and three Euler angles per timestamp, totaling $36$ kinematic variables. Original sampling rates were $179.695$Hz for VTS and $100$Hz for BRS, downsampled to $30$Hz with a Parks-MacClellan FIR low-pass filter ($10$Hz pass-band cutoff, $15$Hz stop-band cutoff).

A full description of data acquisition and prepossessing is described in Supplementary materials \ref{subsec:acquisition_full}.

\subsection{JHU-ISI Gesture and Skill Assessment Working Set (JIGSAWS)}
\label{subsec:JIGSAWS}

In this study, eight surgeons with different skill levels performed five repetitions of three elementary surgical tasks on a benchtop model using the da Vinci Surgical System: suturing, knotting, and needle passing, which are standard components of surgical skills training. The dataset contains stereo video data, kinematic data, and manual annotations of gestures and skills scoring. Consistent with previous research, our work focuses on the suturing task, which involves ten distinct gestures.
The Kinematic data includes data from the two patient-side manipulators (PSMs) and master tool manipulators (MTMs). Each PSM's data includes the cartesian positions, a rotation matrix, linear velocities, angular velocities, and a gripper angle of the manipulator. Here, for each manipulator, we used only three position variables, calculated three Euler angles based on the rotation matrix data, and the gripper angle, in total 14 kinematic variables.
We used the corrected version of the labels, as it was used in \cite{van2020multi}.
As the kinematic data in the JIGSAWS dataset is already provided at 30Hz, we did not filter this data.

\section{Action Segmentation}
Using kinematic data as inputs, we introduce a new, multi-stage network that consists of a temporal convolutional prediction generator in conjunction with a RNN-based refinement stage for the action segmentation task. We refer to RNN as the generic name for several well-known algorithms, including BiLSTM networks and \emph{Bidirectional Gated recurrent units} (BiGRUs).

In action segmentation, given the sequence of vectors including the kinematic data for each timestamp $x_{1:T}=(x_{1},\dots,x_{T})$, the algorithm's goal is to predict one label out of a pre-defined set of classes for each data point. Let  $y_{i}\in \{1,\dots, C\} : \,\, i \in [1,T]$ be the ground truth label for the $i^{th}$ timestamp, and $Y=(y_{1},\dots,y_{T}) \in \Reals^{T}$ be the ground truth sequence, where C is the number of classes and $T$ is the sequence length. The output of the algorithm is a sequence of probabilities vectors $\hat{Y}= (\hat{y}_{1},\dots,\hat{y}_{T}) \in \Reals^{C \times T}$.

\subsection{Multi-Stage Temporal Convolutional Recurrent Networks}
\begin{figure}[ht]
\centering
\includegraphics[width=.6\textwidth]{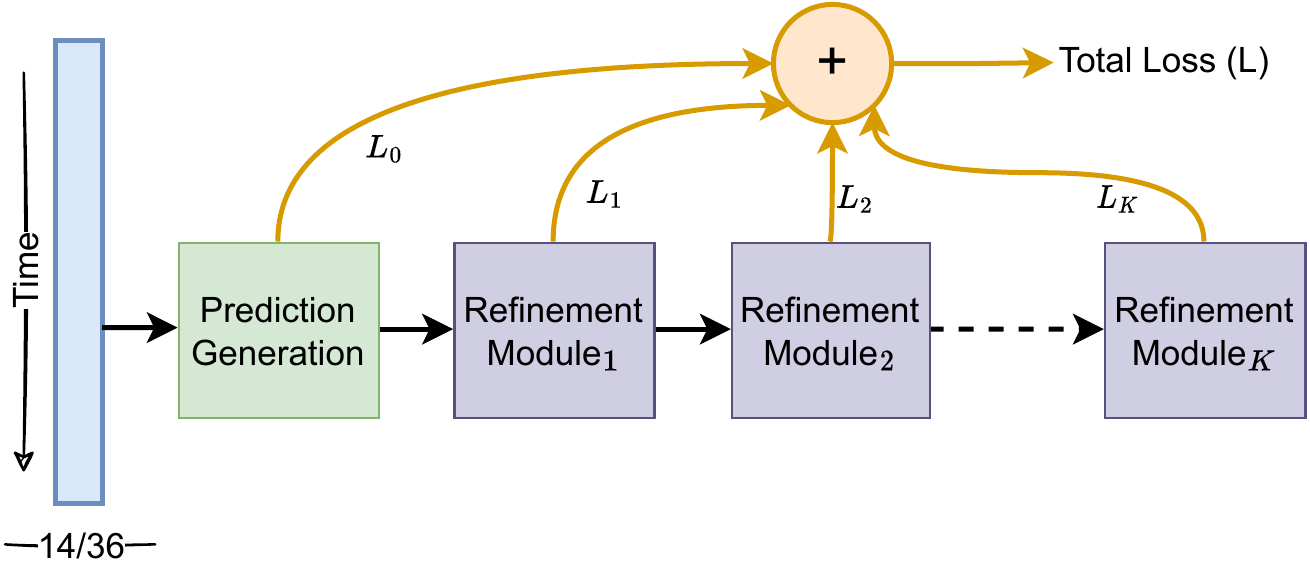}
\caption{General structure of the multi-stage network}
 \label{Figure:MS-framework}
\end{figure}

In the multi-stage framework, there are several models stacked in a sequential manner so that each model's output is used as input for the subsequent model in the chain that refines it. Several tasks, including action segmentation in MS-TCN++, have been found to benefit from the use of this method as a performance enhancer. 
In this chain, the first model serves as a prediction generator, $PG$, and the rest as refinement stages, where $Ref_{j} : j\in \PI$ is the $j^{th}$ refinement stage, as depicted in \Fig \ref{Figure:MS-framework}. Formally, the multi-stage framework is defined as follows:
\begin{align}
    &\hat{Y}_{0} = PG(x_{1:T})\\
    &\hat{Y}_{j} = Ref_{j}(\hat{Y}_{j-1})
\end{align}

Where $\hat{Y}_{0}$ is the output of the prediction generator and $\hat{Y}_{j}: j\in \PI$ is the output of the $j^{th}$ refinement stage.
We would like to point out that in our framework, the input to the refinement stages is just the frame-wise probabilities, without additional information or features from the deeper layers of the model.

\subsubsection{Prediction Generation Module}
We based our prediction generator on the prediction generator of MS-TCN++ with additional prediction heads added inside the stage.
By having these heads, the system can be forced to classify actions based on partial information, with a greater focus placed on the classification data point region, rather than relying solely on the full context obtained in the output of the stage. Hence, we called this method intra-stage regularization (ISR). This method was found to be beneficial to the frame-wise performance of the system. 

We will first discuss the conventional prediction generator as stated in \cite{li2020ms}, and then we will offer our ISR extension.

\begin{figure}[ht]
 \centering
  \includegraphics[width=0.8\columnwidth]{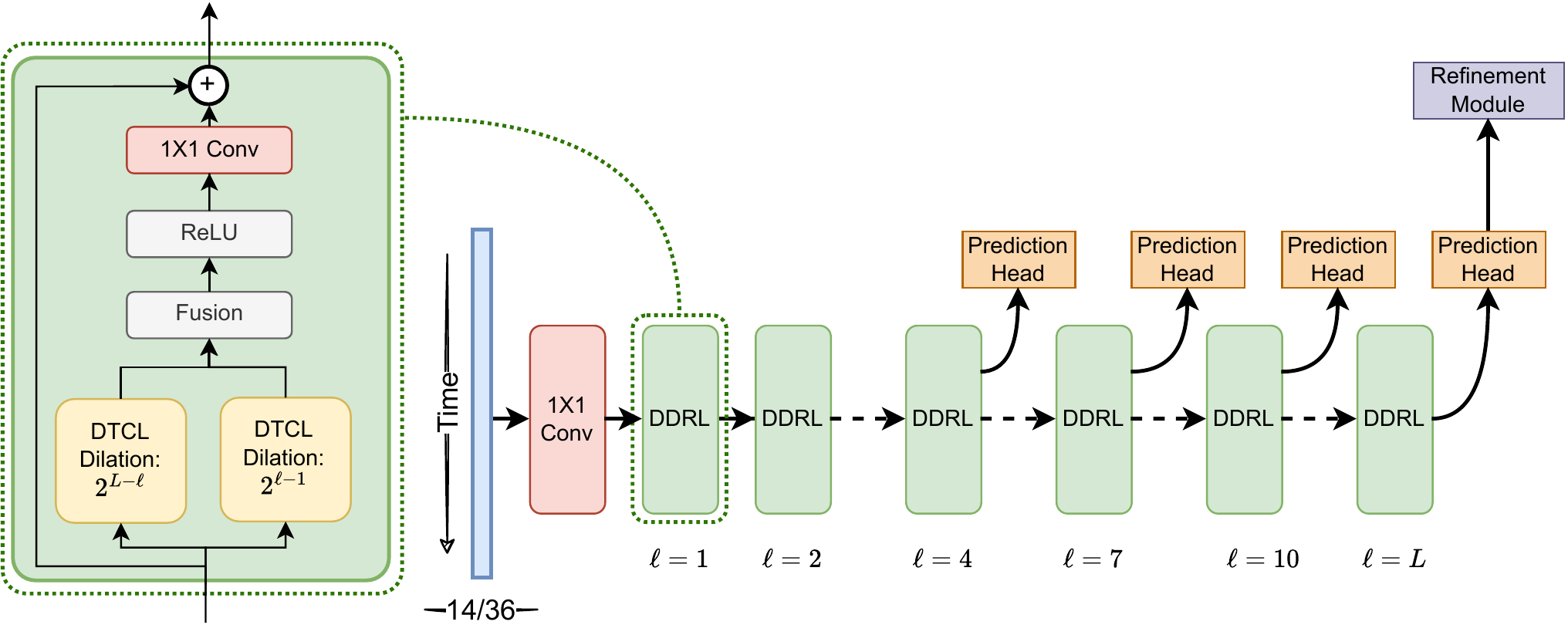}
 \caption{Prediction generator with intra-unit regularization with a close-up view of a dual dilated residual layer (DDRL).}
     \label{Figure:PG}
 \end{figure}

The standard prediction generator consists only of temporal convolutional layers, which allow handling varying input lengths.
To match the input dimensions of the stage with the number of feature maps, the input of the prediction generation unit passes through a $1 \times 1 $ convolutional layer. Then, these features are fed into several \emph{Dual Dilated Residual Layers} (DDRLs), which serve as the prediction generator's core units. In the standard prediction generator in MS-TCN++, as a final step, the output of the last DDRL is passed through a prediction head that includes a $1 \times 1$ convolution to adjust the number of channels to the number of classes.
Let's consider the $\ell^{th}$ DDRL where $\ell \in \{1,2,3 \dots L\}$ and $L$ is the total number of layers.
First, the input of the DDRL is entered into two dilated temporal convolution layers (DTCLs) in parallel, with a kernel size of 3, one with a dilation factor of $\delta^1(\ell)= 2^{\ell - 1}$ and the other with a dilation factor of $\delta^2(\ell)= 2^{L- \ell}$. The dilation factor determines the distance between kernel elements, such that a dilation of 1 means that the kernel is dense. 
Then by concatenating the feature in the channel dimension, the outputs of the two dilated convolutional layers are fused and inserted into a 1D convolutional layer to reduce the number of channels back into the constant number of feature maps. The output passes through a ReLU activation and an additional 1D convolutional layer before the residual connection (see \Fig \ref{Figure:PG}).
Formally, the DDRL can be described as follows:\\
    \begin{align}
    &\hat{H}_{\delta^1(\ell)} = W_{\delta^1} * H_{\ell -1} + b_{\delta^{1}}\\
    &\hat{H}_{\delta^2(\ell)} = W_{\delta^2} * H_{\ell -1} + b_{\delta^{2}}\\
    &\hat{H}_{\ell} = ReLU([\hat{H}_{\delta^1(\ell)}, \hat{H}_{\delta^2(\ell)}])\\
    &H_{\ell} = H_{\ell -1} + W * \hat{H}_{\ell} + b
    \end{align}

Where $H_{\ell}$ is the output of layer $\ell$, $[\cdot,\cdot]$ denotes the concatenation operator and $*$ is the convolution operator,
$W_{\delta^1},W_{\delta^2} \in \Reals^{3 \times D \times D}$ are the weights of temporal dilated convolutions such that $\delta_{1},\delta_{2}$ are their dilation factors, $D$ is the number of feature maps, and 3 is the kernel size. The weights of the $1 \times 1$ convolution are denoted by $W \in \Reals^{1 \times 2D \times D}$, and $b,b_{\delta^{1}},b_{\delta^{2}} \in \Reals^{D}$ are the bias vectors.

\textbf{Prediction generator with intra-stage regularization (ISR):}

As described before, the input and the output of each DDRL have the same dimensions. Thus, each such feature map could, in principle, have been passed to an identical prediction head. In practice, only the output of the final layer is passed to the prediction head.
We refer to this prediction head as the stage's prediction head and to its output as the stage's output. Note that the stage's output is then fed to the refinement stage.

In addition, each feature map from a pre-determined set of layers is passed through a prediction head (see \Fig \ref{Figure:PG}), and contributes equally to the loss.
The core insight driving this idea is that shallower layers concentrate more input vectors around the current time, t, which in turn allows predictions based on lower layers to capture more localized temporal characteristics. In other words, the closer proximity of input vectors to time t in shallower layers enhances the model's ability to capture local temporal patterns for more accurate predictions.
Let $\hat{H}_{ISR}= \{4,7,10\}$ to be a \emph{possible ISR heads indexes set} and
${H}_{ISR}= \{ i \in \hat{H}_{ISR} \mid i < L \}$  to be an \emph{ISR heads indexes set}, namely a set of DDRLs indexes, such that if $i\in H_{ISR}$, then after the $i^{th}$ DDRL in the prediction generator we add a prediction head. Note the stage output head follows layer number $L$ in any case.

The formulation of ISR prediction heads and the stage output head are identical. The prediction heads consist of a $1 \times 1$ convolution over the output features of its DDRL, followed by a softmax activation, formally
\begin{equation}
    O_{t,\ell} = Softmax(W\, h_{\ell,t} + b)
\end{equation}
Where $h_{\ell,t}$ is the output of the $\ell^{th}$ DDRL at time $t$, $W \in \Reals^{C \times D}$ has the weights of $1 \times 1$ convolution and $b \in \Reals^{C}$ is the bias.
$O_{t,\ell}$ represents the class probabilities of the prediction head that follows the $\ell^{th}$ DDRL at time $t$, such that $\ell \in H_{ISR} \cup \{L\} $.
In the case that $\ell = L$ it is satisfied that $O_{t,\ell}= \hat{y}_{t}$, where $\hat{y}_{t}$ contains the stage's output probabilities at time $t$.

\subsubsection{Refinement Stages}
For the refinement stage, we evaluate RNN-based architectures, BiGRU, and BiLSTM. The RNN unit is followed by a linear layer for prediction probabilities generation. 
During training, before entering the prediction head, we apply a dropout.
In our previous study, we found that the optimal frequency for action segmentation of RNN-based networks for kinematic data is significantly distinct from the optimal input frequency of TCN-based networks \cite{goldbraikh2022using}. This fact inspired us to feed the refinement stage with a different frequency than our TCN-based prediction generator.
Hence, the RNN units input is downsampled by factor $k$.
Since our model is evaluated in the full input's resolution, the output is upsampled back into the original frequency.
Down-sampling by a factor of $k$ means that every $k^{th}$ element in the sequence is selected; we mark this operator by $\downarrow k(\cdot)$, namely let $X = (x_1,x_2,\dots, x_N)$ be a sequence of samples $\downarrow k(X) = (x_1,x_{k+1},\dots, x_{N - (N-1)mod k})$, in total there are $N'= \lfloor(N-1)/k\rfloor +1$ elements.
The upsampling operator $\uparrow k(\cdot)$ takes every sample in a sequence and replaces it with $k$ copies of the sample, such that let $X = (x_1,x_2,\dots, x_{N'})$ be a sequence of samples $\uparrow k(X) = (x_{1,1},\dots,x_{1,k},\dots,x_{{N'},1},\dots,x_{{N'},k})$.
To define formally the refinement stage we will use $RNN(\cdot)$ to represent any bidirectional function belonging to the RNN family, such as $BiLSTM(\cdot)$ or $BiGRU(\cdot)$.
Let $\hat{Y}_{j} = (\hat{y}_{j,1},\dots,\hat{y}_{j,T})$ be a probabilities vector obtained from stage number $j \in \N $ which serves as an input to the current refinement stage, and let $k \in \PI$ be a sampling factor.

    \begin{align}
        &\bar{Y}_{j} = \downarrow k(\hat{Y}_{j})\\
        & h_{i} = RNN(\bar{y}_{j,i}) : i \in \{1,\dots,\lfloor(T-1)/k\rfloor +1   \} \label{eq:RNN} \\
        &\bar{y}_{j+1,i} = Softmax(Ah_{i}^{\intercal} + b)\\
        &\hat{Y}_{j+1} = \uparrow k(\bar{Y}_{j+1})
     \end{align}

Where $h_{i} \in \Reals^{1 \times 2Q}$ is the concatenation of forward and reverse hidden states of the last layer in the RNN corresponding to time $t$, and the hidden size of each RNN is $Q$. The weights matrix is $A \in \Reals^{C \times 2Q}$, and $b \in \Reals^{C}$ is a bias vector. We formalized the RNN in the Supplementary Materials \ref{sub_sec: BiRNN formulation}.\\

\subsubsection{Loss Function}
 We based the loss function on the MS-TCN++ loss function, adapting it to our prediction generator with ISR. This loss is defined as the sum of the prediction generation, which consists of the prediction heads losses, and the refinement stages losses.  In total, the loss is $\mathcal{L} = \sum_{\eta} \mathcal{L}_{\eta}$, where $\eta$ represents a prediction head, whether an ISR head or stage's output head. The loss of each prediction head is defined as the weighted sum of the cross-entropy loss and a smoothing loss, formally defined as $\mathcal{L}_{\eta} =-\frac{1}{T} \sum_{t,c} y_{t,c}\log(\hat{y}_{t,c}) + \lambda \cdot \frac{1}{T\cdot C} \sum_{t,c} \Delta_{t.c}^{2}$, where $\lambda$ determines the weight of the smoothing loss. The cross-entropy loss is  $-\frac{1}{T} \sum_{t,c} y_{t,c}  \log(\hat{y}_{t,c}) $, where $\hat{y}_{t,c}$ represents the predicted probability for the class $c$ at time point $t$, $y_{t,c}\in\{0,1\}$ equals one if class c is the ground truth label otherwise, it is zero, and $T$ is the total number of data points, i.e., the procedure length. The smoothing loss is $\frac{1}{T\cdot C} \sum_{t,c} \Delta_{t,c}^{2}$, where $C$ is the total number of classes and $\Delta_{t,c}$ represents the bounded mean error over the procedure-wise log-probabilities. Formally, $\Delta_{t,c}= \min \{\tau, \lvert\log \hat{y}_{t,c} - \log \hat{y}_{t-1,c}\rvert \}$, where $\tau$ is a bounding parameter.

\subsection{Data Augmentations}
In this section, we present two data augmentations: \emph{World frame Rotation} (WFR) to simulate data acquired when the world coordinate system is rotated, as a result, the position and orientation of each sensor in every data point are changed, and the \emph{Hand Inversion} (HI) augmentation where we simulate a work pattern of left-handed surgeons. For each procedure, during the training, each of the augmentations is applied independently in a probabilistic manner with some probability. These two data augmentations improve the generalization ability of our networks.

\subsubsection{World frame Rotation Augmentation}

Kinematic data is obtained using motion sensors that record their pose (location and orientation) at each timestamp. All sensor poses are interrelated within the same coordinate system, requiring consistent rotational operations across all sensors and timestamps for world frame rotation augmentation.

Our networks need orientation data as intrinsic $ZYX$ Euler angles. To apply rotation, the orientation must be transformed into a rotation matrix and then converted back to Euler angles for network input. The electromagnetic tracking system uses multiple sensors with coordinates relative to the transmitter's coordinate system.
 In our two open surgery datasets, there are six sensors on the surgeon's hands. Consider some sensor $s_{i}: i\in [6]$. In the JIGSAWS dataset, there are the left and right PSMs, where the PSM's data is identical to the sensor's data; hence, in this case, we will refer to PSM as a sensor, such that for JIGSAWS $s_{i}: i\in [2]$. The sensor's pose at time $t$ is determined by six degrees of freedom (6 DOF) $s_{i,t}= (x,y,z,e1,e2,e3)$, where $P^{s_{i,t}}_{w}=(x,y,z)$ is the position vector and $E^{s_{i,t}}_{w} = (e1,e2,e3) \in \Reals^3$ is the orientation vector represented by intrinsic $ZYX$ Euler angles, with respect to world frame $w$ that is established with respect to either the electromagnetic sensors coordinate system or the da Vinci coordinate system.
In addition to Euler angles, the orientation of each sensor can be represented by a rotation matrix $R^{s_{i,t}}_{w} \in \Reals^{3 \times 3}$, that represents the rotation from the world frame $w$ to the $s_{i,t}$ sensor's frame. Let $\ROT(E^{s_{i,t}}_{w})= R^{s_{i,t}}_{w}: \Reals^{3} \rightarrow \Reals^{3 \times 3}$ be a representation function from the Euler angles representation to the rotation matrices representation, and  $\ROT^{-1} :  \Reals^{3 \times 3} \rightarrow \Reals^{3}$ be a function from the rotation matrices to Euler angle representation. Note,  $\ROT^{-1}(\cdot)$ function is not a mathematical inverse of $\ROT(\cdot)$, since there is no one-to-one mapping between Euler angles and rotation matrices spaces; as such, these notations represent only the transformation between the two representations. 
Let $w'$ be another world frame with the same origin as $w$; namely, there is no translation between $w$ and $w'$.
The matrix that represents the rotation from the new world frame $w'$ to the sensor's frame $s_{i,t}$ is given by $R^{s_{i,t}}_{w'}= R^{w}_{w'}R^{s_{i,t}}_{w} \in \Reals^{3}$, where $R^{w}_{w'}$ is our \emph{augmentation matrix}.
The new position vector with respect to the new world frame $w'$ is obtained by $P^{s_{i,t}}_{w'}= R^{w}_{w'} \cdot P^{s_{i,t}}_{w} \in \Reals^{3}$.\\
Let $\theta_{max}$ be the maximum allowed rotation angle. First of all, we select uniformly at random three Euler angles $E^{w}_{w'}=(e_1',e_2',e_3')$ such that $e'_{i} \in [-\theta_{max},\theta_{max}]$, where $\theta_{max}$ is a hyperparameter. Then, based on these three Euler angles we calculate the augmentation matrix $R^{w}_{w'}= \ROT(E^{w}_{w'})$.
To change the orientation of the entire procedure, it is required to apply the same augmentation matrix on all six sensors for each time point, such that for each sensor, for each data point we calculate $R^{s_{i,t}}_{w} = \ROT(E^{s_{i,t}}_{w})$. Then we obtain the sensor orientation with respect to the new world frame by $R^{s_{i,t}}_{w'} = R^{w}_{w'} R^{s_{i,t}}_{w}$ and we get back the Euler angle representation by $E_{w'}^{s_{i,t}} = \ROT^{-1}(R^{s_{i,t}}_{w'})$.
The position vector with respect to the new world frame is obtained by $P^{s_{i,t}}_{w'}= R^{w}_{w'} \cdot P^{s_{i,t}}_{w}$.

\subsubsection{Hand Inversion Augmentation}
The \emph{Hand Inversion} (HI) augmentation involves mirroring between left and right hands. In video, this is straightforward with a horizontal flip and label switch. However, sensor data with 3D poses lacks a defined reflection plane, complicating this augmentation.

To develop this augmentation, we will assume that the original axis of the data $z$ is upward. Thus we will define the reflection axis in the $xy$ plane. We will then flip the $xy$ positions across this axis while preserving the values in the $z$ direction. In addition, we will rotate the orientation of all the sensors.

More specifically, in our  open surgery datasets, three sensors were placed on each hand. Hence, we will first calculate for each hand the average location of the three sensors, for each timestamp $t$. Formally, let $\bar{P}_{L,t}=(P^{s_{1,t}}_{w}+P^{s_{2,t}}_{w}+P^{s_{3,t}}_{w})/3$ and $\bar{P}_{R,t}=(P^{s_{4,t}}_{w}+P^{s_{5,t}}_{w}+P^{s_{6,t}}_{w})/3$, where the left hand is denoted by $L$ and the right hand by $R$.
For the JIGSAWS dataset, we will use the left and right PSM's data as is, namely
$\bar{P}_{L,t}= P^{s_{1,t}}_{w}$ and $\bar{P}_{R,t}=P^{s_{2,t}}$. For computational reasons, we first downsample our data points by a factor of $50$.

Next, for each point, we consider only its projection on the $xy$ plane. As such let $\bar{P}_{L,t}^{xy}$ be the projection on the $xy$ plane of $\bar{P}_{L,t}$, and $\bar{P}_{R,t}^{xy}$ be the projection of $\bar{P}_{R,t}$.
Next, we define the reflection axis, which will yield the reflection plane. This axis is the one that separates, in an optimal way, between the left and right-hand data points: $\bar{P}_{L,t}^{xy}$ and $\bar{P}_{R,t}^{xy}$. We will use a geometric interpretation of a linear SVM to calculate this separation.  We then calculate the reflection axis $\chi$ that satisfies the linear equation $\chi: y= mx + b$.
Next, we reflect the original sensors' data across the plane induced by this line, where we assume that $z$ values are preserved.
The reflection matrices are described in \Eq\ref{eq:1}, $Ref_{2D}$ is derived from the angle $\phi$ of the line $\chi$ with the x-axis, such that $\phi= \arctan(m)$. 
Since the $z$ values are preserved, our $3D$ reflection matrix is obtained by a trivial extension.
\begin{equation}
\label{eq:1}
Ref_{2D}(\phi)= \begin{bmatrix}
\cos{2\phi} & \sin{2\phi} \\
\sin{2\phi} & -\cos{2\phi} 
\end{bmatrix} , Ref_{3D}(\phi)= \begin{bmatrix}
\cos{2\phi} & \sin{2\phi} & 0 \\
\sin{2\phi} & -\cos{2\phi} & 0 \\
0 & 0 & 1 
\end{bmatrix}
\end{equation}

Since our reflection line $\chi$ has a bias element $b$, for every sensor position $P^{s_{i,t}}_{w}=(x,y,z)$ we will subtract the bias, resulting $P^{s_{i,t}}_{\bar{w}} = (x,y - b,z)$. Where $\bar{w}$ is the world frame after this translation.
The next step is to apply the reflection matrix on each position vector. Let us denote the reflected world frame by $w_{Ref'}$, then $P^{s_{i,t}}_{w_{Ref'}} = Ref_{3D}(\phi) \cdot P^{s_{i,t}}_{\bar{w}} = (x_{Ref'},y_{Ref'},z_{Ref'}) \in \Reals^{3}$; notice that $Ref_{3D}(\phi) \in \Reals^{3 \times 3}$ is a matrix, and  $P^{s_{i,t}}_{\bar{w}} \in \Reals^{3}$ is a vector. To calculate the final portion vector we will add back the bias: $P^{s_{i,t}}_{w_{Ref}} = (x_{Ref'},y_{Ref'} + b,z_{Ref'})$, where $w_{Ref}$ is the final reflected world frame.

Position augmentation will be followed by augmentation of the sensor orientation. For each sensor and timestamp, we will represent the orientation by its rotation matrix, namely for each $E^{s_{i,t}}_{w}$ we will obtain $R^{s_{i,t}}_{w} = \ROT(E^{s_{i,t}}_{w}) \in \Reals^{3 \times 3}$. Then, we will obtain the reflected orientation in rotation matrix representation by multiplying this matrix by our reflection matrix, such that $R^{s_{i,t}}_{w_{Ref}}= R^{s_{i,t}}_{w} \cdot Ref_{3D}(\phi)$. Then, we calculate the reflected Euler angles representation by $E_{w_{Ref}}^{s_{i,t}} = \ROT^{-1}(R^{s_{i,t}}_{w_{Ref}})$. Finally, we exchanged between the surgeon's right hand and the left hand, such that for our open surgery datasets, sensor 1 was switched with sensor 4, sensor 2 was switched with sensor 5, and sensor 3 was exchanged with sensor 6. For JIGSAWS datasets the left PSM was switched with the right PSM.

  \begin{figure*}[ht]
 \centering
\includegraphics[width=1.05\textwidth]{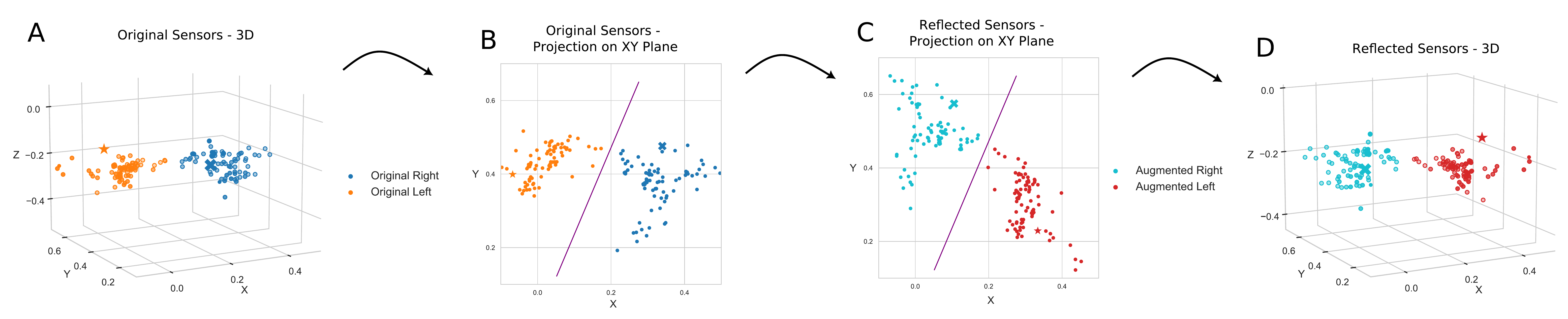}
 \caption{Sample points from two sensors, one on each hand, illustrating the Hand Inversion augmentation. The star- and cross-shaped points emphasize the transformations. The reflection plane given by the SVM output is represented by the purple line. \textbf{A} shows the original points in 3D, and \textbf{B} in the XY plane projection. The flipped points are shown in the XY plane in \textbf{C} and in 3D in \textbf{D}. Points in dark blue (orange) and light blue (red) represent points from the right (left) hand before and after augmentation respectively. Augmentation of orientation is not displayed.
 }
 \label{Figure:flip-aug}
 \end{figure*}

\section{EXPERIMENTS}
\subsection{Data Flow}
\begin{figure*}[ht]
 \centering
\includegraphics[width=\textwidth]{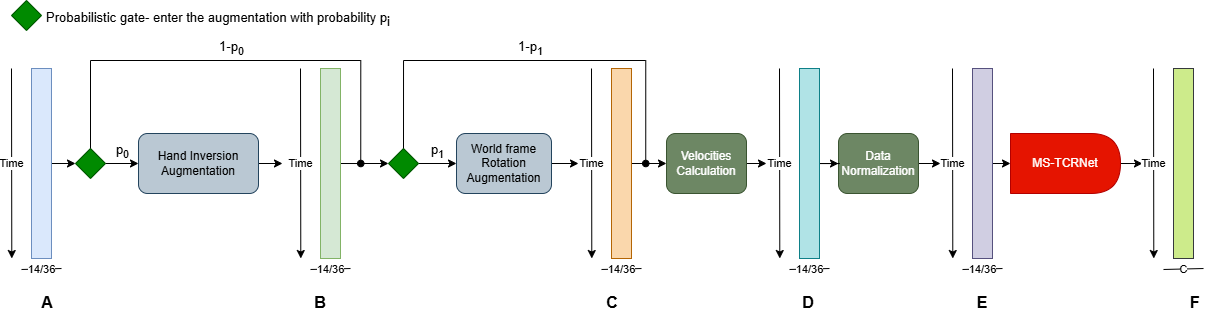} 
    \caption{The complete data flow in the training process is depicted here. A-E represents the input at different preprocessing stages, while F is the output, with dimensions equal to the number of classes multiplied by the number of time samples. A- shows the raw input containing the position and orientation of each sensor, B- displays the position and orientation after Hand Inversion augmentation, C- presents the position and orientation after World Frame Rotation augmentation, D- contains the calculated linear and angular velocities, and E- shows the normalized velocities.
 }
    \label{fig:data_flow}
\end{figure*}

Based on previous studies, velocities are the most beneficial input to an algorithm for analyzing kinematic data.
Our augmentations must, however, be applied at absolute locations and orientations. Hence, as raw data, we used locations and Euler angles, then applied our augmentations, and finally, before feeding the input into the algorithm, we calculated the velocities and normalized the data. 
Kinematic data in the JIGSAWS dataset do not include Euler angles, but rather rotation matrices. Based on these rotation matrices, we calculated the intrinsic Euler angles as a preprocessing step.
The complete data flow is illustrated in \ref{fig:data_flow}.

\subsubsection{Velocities Calculation}
For all three datasets, based on the Euler angles and positions, for each coordinate, the differences between every two adjacent timestamps were calculated; thus, the input was the calculated linear and angular velocities.
  
\subsubsection{Normalization}
For each coordinate $j$ in the input sequence, the data were normalized using the Standard score, formally defined as
$\hat{\xi}^{j}_{1:T}=\frac{\xi^{j}_{1:T}-\bar{\xi}^{j}}{S^{j}}$, where $\xi^{j}_{1:T} \in \Reals^{T}$ is a vector representing the non-normalized input data values, $\hat{\xi}^{j}_{1:T} \in \Reals^{T}$ the normalized data values, $\bar{\xi}^{j}$ the mean, and $S$ the standard deviations of the elements of $\xi^{j}_{1:T} \in \Reals^{T}$, where $\bar{\xi}^{j}$ and $S$ are calculated separately for each procedure.

\subsection{Evaluation Method}
\label{sec:Evaluation method}

We used two groups of metrics for model evaluation: frame-wise and segmentation metrics. Frame-wise metrics include Accuracy and Macro F1, suitable for imbalanced classes. Segmentation metrics include Edit distance\footnote{Marked as Edit* in \cite{van2021gesture}}, normalized by the maximum length between ground truth and prediction, and F1@k with $k \in {10, 25, 50}$ \cite{lea2017temporal}. 

All models trained on VTS and BRS datasets underwent 5-fold cross-validation. Sequences were split by participant to ensure each participant's procedures stayed within the same fold. The $i^{th}$ fold's sequences constituted the test set, with the remaining sequences split into training and validation sets. The validation set, containing procedures from the $(i+1) \mod 5$ fold, was used for hyperparameter tuning and stopping criteria, ensuring disjoint sets at the participant level. In VTS, three participants (12 procedures) and in BRS, five participants were selected from the $(i+1) \mod 5$ fold for validation. JIGSAWS used leave-one-user-out cross-validation \cite{gao2014jhu}.
To ensure a fair comparison with \cite{goldbraikh2022using} on VTS, we used their dataset and settings. Additionally, we separately evaluated the HI augmentation on a left-handed surgeon, which was excluded in \cite{goldbraikh2022using}.

To ensure the stability of the networks, in addition to cross-validation, multiple random seeds were used. Eight different seeds were used for the VTS dataset and six for the BRS dataset. The reported results for each metric are the mean and the standard deviation across all procedures of the dataset and all random seeds. In the case of  the VTS dataset, there are $96 \,\, procedures \times 8 \,\, seeds$, and in the case of BRS dataset, there are $52 \,\, procedures \times 6 \,\, seeds$. In both datasets, networks were trained for 40 epochs, and extended to 80 epochs with Hand Inversion augmentation.

    For JIGSAWS, a single predetermined seed was used, consistent with prior studies. Networks trained for 60 epochs, or 90 with Hand Inversion augmentation, ensuring convergence. The reported results are from the final epoch. Additionally, in JIGSAWS training, a scheduler halved the learning rate if loss stagnated for 3 consecutive epochs.

\subsection{Hyperparameter Tuning}
\label{sec: Hyperparameter_tuning}
    The hyperparameter optimization was executed only on the VTS dataset on both our L-MS-TCRNet and G-MS-TCRNet networks using the Optuna environment \cite{optuna_2019} with the TPESampler sampler for 300 trials.
    Full description of the hyperparameter tuning is described in Supplementary Materials \ref{sub_sec:Hyperparameter_Tuning_full}.

\subsection{Ablation Study}
In this section, we investigate the performance of our networks by replacing certain components to understand the contribution of each component to the overall system. We started with MS-TCN++ adapted to kinematic data as proposed in \cite{goldbraikh2022using}, then we replaced every element in the architecture and measured the impact of this change. Note that we can use this method since we preserved the multi-stage framework of the architecture and replaced one component with another one that plays the same role.
First of all, we replaced the standard prediction generator with our prediction generator which includes the ISR. It was found, as expected, that this step contributes to frame-wise metrics, but unfortunately at the expense of segmental metrics. Next, we showed that replacing the refinement stages with our RNN-based refinements improves the segmental performance of the networks; this result is also in line with our assumptions during the development of these refinement stages. Finally, we showed that procedure-wise normalization contributes to the performance of our proposed networks. Note that in \cite{goldbraikh2022using}, the mean and the standard deviation were calculated a priori based on the whole train set for each fold individually, instead of for each sequence separately.
\Table \ref{tab:VTS-ablation} summarizes the impact of each modification on the architecture. We found that each of the modifications contributes to the overall system in at least one type of metric.

\begin{table}[ht]
\centering
\caption{Ablation study of the proposed architectures on the gesture recognition task of the VTS dataset - without data augmentation. * represents experiments using the normalization method from \cite{goldbraikh2022using}.}
\label{tab:VTS-ablation}
\scalebox{0.7}{
\begin{tabular}{@{}lcccccc@{}}
\toprule
\textbf{VTS} & \textbf{F1-macro} & \textbf{Acc} & \textbf{Edit} & \textbf{F1@10} & \textbf{F1@25} & \textbf{F1@50} \\ \midrule
\textbf{MS-TCN++* \cite{goldbraikh2022using}} & 78.9 & 82.4 & 86.3 & 89.3 & 85.8 & 71.1 \\
\textbf{MS TCN++ ISR*} & 79.8 & 83.3 & 81.7 & 86.2 & 82.8 & 68.6 \\
\textbf{L-MS-TCRNet*} & 79.5 & 83 & 88.2 & 91 & 87.7 & 73.2 \\
\textbf{G-MS-TCRNet*} & 79.9 & 83.1 & \textbf{89.3} & 91.5 & 87.8 & 73.4 \\ \midrule
\textbf{L-MS-TCRNet} & 80.4 & 83.7 & 87.7 & 91.1 & 88.2 & 75.1 \\
\textbf{G-MS-TCRNet} & \textbf{80.9} & \textbf{84.1} & 89.0 & \textbf{91.7} & \textbf{88.7} & \textbf{75.5} \\ \bottomrule
\end{tabular}}%

\end{table}

\subsection{BRS Dataset Baseline Establishing}
Since the BRS dataset is introduced here for the first time for action segmentation without previously reported results, we established a baseline by training an MS-TCN++ model with the procedure-wise normalization method. We then compared this baseline to the G-MS-TCRNet and L-MS-TCRNet models that we propose. The results are presented in \Table \ref{Table:Baseline_BRS}. We found that the standard MS-TCN++ has significant difficulty with segmental metrics on this dataset, but both our architectures performed significantly better. The performance gap between MS-TCN++ and our algorithms in the BRS dataset is larger, highlighting the generalization advantages of our approach.

\begin{table}[ht]
\centering
\caption{Comparison between MS-TCN++, G-MS-TCRNet, and L-MS-TCRNet on the BRS dataset.}
\label{Table:Baseline_BRS}
\scalebox{0.7}{
\begin{tabular}{@{}lcccccc@{}}
\toprule
\textbf{BRS} & \textbf{F1-macro} & \textbf{Acc} & \textbf{Edit} & \textbf{F1@10} & \textbf{F1@25} & \textbf{F1@50} \\ \midrule
\textbf{MS-TCN++} & 64.4 & 76.7 & 45.8 & 52.7 & 49.4 & 37.9 \\
\textbf{G-MS-TCRNet} & 64.8 & 77.4 & \textbf{75.8} & \textbf{77.4} & 72.9 & \textbf{57.6} \\
\textbf{L-MS-TCRNet} & \textbf{65.9} & \textbf{77.8} & 75.1 & 77.3 & \textbf{73.1} & 57.1 \\ \bottomrule
\end{tabular}%
}
\end{table}

\subsection{World Frame Rotation}
In this section, we present the effect of our WFR augmentation. We performed several experiments on each of the datasets. 
In the first experiment, we predefined that $\theta_{max}=7\degree$ and evaluated the effect of the probability of activating the augmentation. This experiment was performed on the VTS and BRS datasets. The results of this experiment are presented in \Table \ref{Table:VTS_BRS_Rotation}.
In general, we observe that the effect of this augmentation is noticeable on the BRS dataset but almost negligible on the VTS dataset. The difference in the effect of this augmentation on both datasets can probably be explained by the different methods by which the data were collected. While in VTS, data collection was performed only at one simulation station; in BRS there were eight stations. This might have resulted in some variance between the positioning of the transmitters at different stations. Note that these transmitters define the coordinate frame. Therefore, the ability to generalize the world frame angle contributes to the algorithm's performance on this type of data.
For both models on the BRS dataset, the algorithm performed best when the augmentation was activated for each sequence with a probability of one. Note that during evaluation, we do not apply augmentations.

\begin{table}[ht]

\centering
\caption{WFR augmentation evaluation on the VTS and BRS datasets. $\theta_{max}$ is fixed at $7\degree$ and the augmentation is applied at different probabilities.}
\label{Table:VTS_BRS_Rotation}
\scalebox{0.7}{
\begin{tabular}{@{}llcccccc@{}}
\toprule
\textbf{VTS} & \textbf{prob} & \textbf{F1-macro} & \textbf{Acc} & \textbf{Edit} & \textbf{F1@10} & \textbf{F1@25} & \textbf{F1@50} \\ \midrule
\multirow{3}{*}{\textbf{G-MS-TCRNet}} & \textbf{0} & \textbf{80.9} & 84.1 & \textbf{89.0} & 91.7 & 88.7 & 75.5 \\
 & \textbf{0.5} & 80.7 & 83.9 & \textbf{89.0} & \textbf{91.8} & 88.7 & 75.4 \\
 & \textbf{1} & 80.2 & 83.5 & 88.4 & 91.4 & 88.6 & 75.1 \\ \midrule
\multirow{3}{*}{\textbf{L-MS-TCRNet}} & \textbf{0} & 80.4 & 83.7 & 87.7 & 91.1 & 88.2 & 75.1 \\
 & \textbf{0.5} & 80.8 & \textbf{84.2} & 88.6 & 91.7 & \textbf{89.2} & \textbf{76.1} \\
 & \textbf{1} & 80.0 & 83.4 & 88.4 & 91.5 & 88.7 & 75.2 \\ \midrule
\textbf{BRS} & \textbf{prob} & \textbf{F1-macro} & \textbf{Acc} & \textbf{Edit} & \textbf{F1@10} & \textbf{F1@25} & \textbf{F1@50} \\ \midrule
\multirow{3}{*}{\textbf{G-MS-TCRNet}} & \textbf{0} & 64.8 & 77.4 & 75.8 & 77.4 & 72.9 & 57.6 \\
 & \textbf{0.5} & 67.4 & 78.8 & 77.0 & 79.2 & 75.1 & \textbf{59.4} \\
 & \textbf{1} & 68.0 & 78.7 & \textbf{77.7} & \textbf{79.4} & 75.0 & 59.3 \\ \midrule
\multirow{3}{*}{\textbf{L-MS-TCRNet}} & \textbf{0} & 65.9 & 77.8 & 75.1 & 77.3 & 73.1 & 57.1 \\
 & \textbf{0.5} & 67.1 & 78.5 & 76.8 & 78.5 & 74.3 & 58.4 \\
 & \textbf{1} & \textbf{68.5} & \textbf{79.2} & 77.3 & 79.2 & \textbf{75.2} & 59.3 \\ \bottomrule
\end{tabular}%
}
\end{table}

Next, we evaluate the effect of $\theta_{max}$ on the performance of our networks, on the BRS dataset. We predetermined that the world frame rotation will be applied with a probability of one. We examine $\theta_{max} \in \{0\degree,7\degree,15\degree\}$. The results are reported in \Table \ref{Table:Rotation_theta_max_exp}.
We found that both algorithms performed best with $\theta_{max} = 7 \degree$. In addition, note that besides the effect on the mean performance values, the standard deviations are also reduced by this augmentation. This is suggestive of the improvement in the algorithm's robustness.\\

\begin{table}[ht]
\centering
\caption{WFR augmentation evaluation on the BRS dataset, with varying degrees of $\theta_{max}$. The augmentation was applied with a probability of one.}
\label{Table:Rotation_theta_max_exp}
\scalebox{0.7}{
\begin{tabular}{@{}lcccccc@{}}
\toprule
\textbf{\textbf{$\mathbf{\theta_{max}}$}} & \textbf{F1-macro} & \textbf{Acc} & \textbf{Edit} & \textbf{F1@10} & \textbf{F1@25} & \textbf{F1@50} \\ \midrule
\multicolumn{7}{c}{\textbf{G-MS-TCRNet}} \\ \midrule
\textbf{$\mathbf{0 \degree}$} & $64.8 \pm   13.5$ & $77.3 \pm 8.7$ & $75.8 \pm 9.9$ & $77.4 \pm 10.1$ & $72.9 \pm 11.7$ & $57.6 \pm 14.6$ \\
\textbf{$\mathbf{7 \degree}$} & $68.0 \pm   12.9$ & $78.7 \pm 8.0$ & $77.7 \pm 10.0$ & $79.4 \pm 9.9$ & $75.0 \pm 12.2$ & $59.3 \pm 14.6$ \\
\textbf{$\mathbf{15 \degree}$} & $67.8 \pm 12.0$ & $79.0 \pm 7.5$ & $77.7 \pm 10.1$ & $79.1 \pm 9.6$ & $75.3 \pm 11.2$ & $60.0 \pm 14.6$ \\ \midrule
\multicolumn{7}{c}{\textbf{L-MS-TCRNet}} \\ \midrule
\textbf{$\mathbf{0 \degree}$} & $65.9 \pm   14.1$ & $77.8 \pm 8.9$ & $75.1 \pm 10.7$ & $77.3 \pm 11.1$ & $73.1 \pm 12.5$ & $57.1 \pm 15.1$ \\
\textbf{$\mathbf{7 \degree}$} & $68.5 \pm   13.1$ & $79.2 \pm 8.5$ & $77.3 \pm 10.0$ & $79.2 \pm 10.6$ & $75.2 \pm 12.4$ & $59.3 \pm 14.7$ \\
\textbf{$\mathbf{15 \degree}$} & $67.8 \pm 13.3$ & $78.9 \pm 8.5$ & $76.9 \pm 10.1$ & $79.1 \pm 10.2$ & $75.3 \pm 12.0$ & $59.9 \pm 14.9$ \\ \bottomrule
\end{tabular}%
}
\end{table}

\subsection{Hand Inversion Effect}

We analyze the effect of HI augmentation on the VTS and BRS datasets but not on the JIGSAWS dataset, as it lacks left-handed surgeons. Goldbraikh et al. \cite{goldbraikh2022using} excluded the single left-handed surgeon in the VTS data, so we evaluate our augmentation on this data separately for direct comparison. The augmentation, applied with a 0.5 probability, creates a pseudo-balanced dataset between left and right-handed surgeons. This is crucial due to the low representation of left-handed surgeons in clinics and datasets, addressing the poor performance of deep learning systems for this group.\\ 

First, we examine the effect of HI augmentation on the left-handed surgeon excluded from the VTS dataset in \cite{goldbraikh2022using}. The left-handed surgeon's four procedures served as a test set for all folds, while the train and validation sets remained standard as defined in \Sect \ref{sec:Evaluation method}. The results are presented in \Table \ref{Table:horizontal-flip-VTS}.
Both networks struggled to generalize to left-handed surgeons without this augmentation. With the augmentation, we achieved performance comparable to that of our regular test set of right-handed surgeons.

\begin{table}[ht]
\centering
\caption{HI evaluation on the left-handed surgeon from the VTS dataset which was excluded from the VTS dataset in \cite{goldbraikh2022using}.}
\label{Table:horizontal-flip-VTS}
\scalebox{0.7}{
\begin{tabular}{@{}llcccccc@{}}
\toprule
VTS & \textbf{HI} & \textbf{F1-macro} & \textbf{Acc} & \textbf{Edit} & \textbf{F1@10} & \textbf{F1@25} & \textbf{F1@50} \\ \midrule
\multirow{2}{*}{\textbf{G-MS-TCRNet}} &     & 26.7 & 45.9 & 34.2 & 32.8 & 25.2 & 12.7 \\
 & \checkmark & \textbf{79.9} & \textbf{84.3} & \textbf{85.0} & \textbf{90.1} & \textbf{88.1} & \textbf{77.8} \\ \midrule
\multirow{2}{*}{\textbf{L-MS-TCRNet}} &     & 32.1 & 48.5 & 43.0 & 39.1 & 31.1 & 14.3 \\
 & \checkmark & 77.6 & 83.0 & 81.6 & 87.2 & 85.3 & 72.5 \\ \bottomrule
\end{tabular}%
}
\end{table}

Next, we examine the effect of Hand Inversion (HI) and its combination with world frame rotation augmentation ($\theta_{max} = 7\degree$, activation probability 1) on our networks. This experiment uses the standard test sets, including three left-handed surgeons, making the results comparable to previous sections on the BRS dataset. HI affects training convergence time, so we trained for 80 epochs. The results are presented in \Table \ref{Table:Rotation_theta_max_exp_wfr}.

For both networks, using this augmentation increased mean performance and reduced standard deviation. L-MS-TCRNet showed further improvement when HI was combined with WFR augmentations during training, resulting in the best performance on this dataset.

\begin{figure}[ht]
 \centering
\includegraphics[width=1\columnwidth]{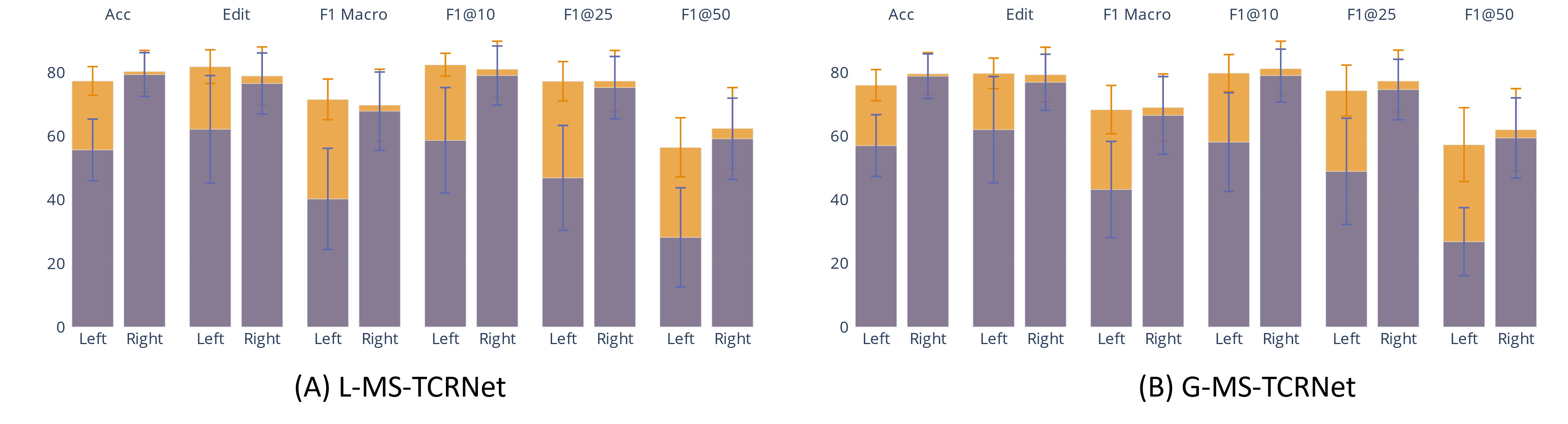}
 \caption{Surgeons' performance by handedness. Purple: without HI; Orange: with HI augmentation. Evaluated A- L-MS-TCRNet and B- G-MS-TCRNet models on BRS dataset. Error bars show standard deviation.}
 \label{Figure:flip-aug-results}
 \end{figure}

Finally, we analyzed the effect of our HI augmentation conditioned on the surgeon's handedness, as shown in \Fig \ref{Figure:flip-aug-results}.
The results are similar to those observed on the VTS dataset in \Table \ref{Table:horizontal-flip-VTS}, where the augmentation significantly impacted left-handed surgeons, leading to comparable performance between left- and right-handed surgeons. Conversely, there was only a minor effect on right-handed surgeons. Thus, augmentation enhances algorithm robustness across both handedness types, contributing to decreased standard deviations.

\begin{table}[ht]
\centering
\caption{The effect of HI and HI+ WFR together. The HI augmentation was applied with a probability of 0.5; for WFR, $\theta_{max}$ is fixed at $7\degree$ and the augmentation was applied with a probability of one.}

\label{Table:Rotation_theta_max_exp_wfr}
\scalebox{0.7}{
\begin{tabular}{@{}lcccccc@{}}
\toprule
\textbf{BRS} & \textbf{F1-macro} & \textbf{Acc} & \textbf{Edit} & \textbf{F1@10} & \textbf{F1@25} & \textbf{F1@50} \\ \midrule
\multicolumn{7}{c}{\textbf{G-MS-TCRNet}} \\ \midrule
\textbf{Baseline} & $64.8 \pm 13.5$ & $77.4 \pm 8.7$ & $75.8 \pm 9.9$ & $77.4 \pm 10.1$ & $72.9 \pm 11.7$ & $57.6 \pm 14.7$ \\
\textbf{HI} & $68.4 \pm 10.6$ & $79.2 \pm 6.6$ & $\mathbf{78.9 \pm 8.5}$ & $80.7 \pm 8.5$ & $76.8 \pm 9.6$ & $61.7 \pm 13.0$ \\
\textbf{HI + WFR} & $67.7 \pm 11.0$ & $78.9 \pm 6.6$ & $78.7 \pm 8.1$ & $80.6 \pm 8.3$ & $76.8 \pm 9.8$ & $61.5 \pm 13.2$ \\ \midrule
\multicolumn{7}{c}{\textbf{L-MS-TCRNet}} \\ \midrule
\textbf{Baseline} & $65.9 \pm   14.1$ & $77.8 \pm 8.9$ & $75.1 \pm   10.7$ & $77.3 \pm   11.1$ & $73.1 \pm 12.5$ & $57.1 \pm 15.1$ \\
\textbf{HI} & $69.4 \pm 11.0$ & $80.0 \pm 6.4$ & $78.7 \pm 9.0$ & $80.7 \pm 8.6$ & $77.1 \pm 9.3$ & $61.8 \pm 12.8$ \\
\textbf{HI + WFR} & $\mathbf{70.0 \pm 10.8}$ & $\mathbf{80.5 \pm 5.9}$ & $78.4 \pm 9.5$ & $\mathbf{81.1 \pm 8.6}$ & $\mathbf{77.5 \pm 9.3}$ & $\mathbf{62.5 \pm 12.2}$ \\ \bottomrule
\end{tabular}%
}
\end{table}

\subsection{JIGSAWS Evaluation}

This experiment evaluates the impact of augmentations on our algorithms' performance on the JIGSAWS dataset. We compare performance without augmentation, with WFR, and with both WFR and HI. WFR used $\theta_{max}=7$ with a probability of 1, and HI was applied with a probability of 0.5. The results are in \Table \ref{Table:JIGSAWS-kin-exp}.

Generally, G-MS-TCRNet outperforms L-MS-TCRNet. Augmentations improve baseline performance for both networks, except HI on L-MS-TCRNet, which shows no effect. Combining both augmentations on G-MS-TCRNet yields the best results. Despite the lack of left-handed surgeons in the dataset, HI augmentation proves highly effective, likely due to enhanced generalization during training.

\begin{table}[ht]
\centering
\caption{The effect of augmentations in evaluation on the JIGSAWS dataset. Where WFR was used, $\theta_{max}$ was fixed at $7 \degree$, and applying the augmentation was with a probability of one. When HI was used, it was applied with a probability of 0.5 }
 
\label{Table:JIGSAWS-kin-exp}
\scalebox{0.7}{
\begin{tabular}{@{}lllllll@{}}
\toprule
\multirow{2}{*}{\textbf{JIGSAWS}} & \multicolumn{1}{c}{\multirow{2}{*}{\textbf{F1-macro}}} & \multicolumn{1}{c}{\multirow{2}{*}{\textbf{Acc}}} & \multicolumn{1}{c}{\multirow{2}{*}{\textbf{Edit}}} & \multicolumn{1}{c}{\multirow{2}{*}{\textbf{F1@10}}} & \multicolumn{1}{c}{\multirow{2}{*}{\textbf{F1@25}}} & \multicolumn{1}{c}{\multirow{2}{*}{\textbf{F1@50}}} \\
                                  & \multicolumn{1}{c}{}                                   & \multicolumn{1}{c}{}                              & \multicolumn{1}{c}{}                               & \multicolumn{1}{c}{}                                & \multicolumn{1}{c}{}                                & \multicolumn{1}{c}{}                                \\ \midrule
\multicolumn{7}{c}{\textbf{L-MS-TCRNet}}                                                                                                                                                                                                                                                                                                                           \\ \midrule
\textbf{Baseline}            & $77.7 \pm 13.7$                                        & $84.2 \pm 10.7$                                   & $88.2 \pm 13.8$                                    & $91.6 \pm 10.3$                                     & $90.8 \pm 11.4$                                     & $85.3 \pm 15.0$                                     \\
\textbf{WFR}        & $78.5 \pm 13.0$                                        & $84.8 \pm 10.0$                                   & $86.9 \pm 13.5$                                    & $91.0 \pm 10.1$                                     & $90.4 \pm 11.3$                                     & $84.0 \pm 15.3$                                     \\
\textbf{WFR+HI}    & $78.1 \pm 12.5$                                        & $84.6 \pm 9.2$                                    & $87.9 \pm 12.8$                                    & $91.7 \pm 9.5$                                      & $90.4 \pm 11.0$                                     & $83.7 \pm 15.7$                                     \\ \midrule
\multicolumn{7}{c}{\textbf{G-MS-TCRNet}}                                                                                                                                                                                                                                                                                                                            \\ \midrule
\textbf{Baseline}             & $79.2 \pm 12.2$                                        & $85.0 \pm 10.1$                                   & $87.7 \pm 10.2$                                    & $91.4 \pm 8.7$                                      & $90.6 \pm 9.9$                                      & $85.1 \pm 14.0$                                     \\
\textbf{WFR}         & $80.0 \pm 11.9$                                        & $85.5 \pm 9.0$                                    & $90.1 \pm 10.4$                                    & $93.3 \pm 8.0$                                      & $92.6 \pm 8.8$                                      & $86.4 \pm 13.5$                                     \\
\textbf{WFR+HI}      & $\mathbf{81.8 \pm 11.2}$                               & $\mathbf{86.4 \pm 7.3}$                           & $\mathbf{90.5 \pm 11.4}$                           & $\mathbf{94.1 \pm 7.5}$                             & $\mathbf{93.6 \pm 8.1}$                             & $\mathbf{87.0 \pm 13.1}$                            \\ \bottomrule
\end{tabular}

}
\end{table}

\subsection{Comparison with The State-Of-The-Art}
We compare our best algorithms with previous state-of-the-art (SOTA) on the VTS and JIGSAWS datasets, as shown in \Table \ref{Table:VTS_JIGSAWS_SOTA}.

For the VTS dataset, both our algorithms outperform those in \cite{goldbraikh2022using}, including the Multi-task (MT) versions. G-MS-TCRNet sets a new benchmark on this dataset.
On the VTS dataset, both our algorithms surpass those in \cite{goldbraikh2022using}, including the Multi-task (MT) versions, with G-MS-TCRNet setting a new benchmark.

For the JIGSAWS dataset, G-MS-TCRNet with WFR-HI augmentation achieves new SOTA results across all kinematic modality metrics. Note, some networks compared used original JIGSAWS labels corrected in \cite{van2020multi} since 2020.

Using G-MS-TCRNet with WFR+HI, accuracy improves by 0.9 and edit score by 2. Previous SOTA results varied by metric, but our single network significantly advances over prior benchmarks.

\begin{table}[ht]
\centering

\caption{Comparison to state-of-the-art results on the VTS and JIGSAWS datasets, for models trained on kinematic data only. Models marked with an * are based on the original JIGSAWS labels.}
\label{Table:VTS_JIGSAWS_SOTA}
\scalebox{0.7}{
\begin{tabular}{@{}llccccc@{}}
\toprule
\textbf{VTS}                                            & \multicolumn{1}{c}{\textbf{F1-macro}} & \textbf{Acc}            & \textbf{Edit}           & \textbf{F1@10}          & \textbf{F1@25}          & \textbf{F1@50}           \\ \midrule
BiGRU \cite{goldbraikh2022using}       & $78.2 \pm 8.8$                        & $82.2 \pm 7.3$          & $84.9 \pm 7.7$          & $88.0 \pm 7.0$          & $83.8 \pm 10.2$         & $68.9 \pm 18.3$          \\
BiLSTM \cite{goldbraikh2022using}       & $77.1 \pm 9.3$                        & $81.3 \pm 7.5$          & $84.7 \pm 8.2$          & $88.1 \pm 7.3$          & $83.7 \pm 10.4$         & $68.1 \pm 18.7$          \\
MS-TCN++ \cite{goldbraikh2022using}     & $78.9 \pm 8.5$                        & $82.4 \pm 7.0$          & $86.3 \pm 8.4$          & $89.3 \pm 7.0$          & $85.8 \pm 9.8$          & $71.1 \pm 17.9$          \\
MT BiGRU \cite{goldbraikh2022using}     & $78.2 \pm 8.7$                        & $82.2 \pm 7.0$          & $85.4 \pm 7.4$          & $88.1 \pm 6.8$          & $83.9 \pm 9.8$          & $69.0 \pm 18.1$          \\
MT BiLSTM \cite{goldbraikh2022using}   & $75.7 \pm 9.3$                        & $79.9 \pm 7.6$          & $83.3 \pm 9.0$          & $86.4 \pm 8.3$          & $81.9 \pm 11.4$         & $65.9 \pm 18.8$          \\
MT MS-TCN++ \cite{goldbraikh2022using}  & $78.5 \pm 8.2$                        & $82.4 \pm 6.6$          & $86.0 \pm 8.5$          & $89.1 \pm 7.5$          & $85.8 \pm 10.0$         & $71.4 \pm 17.5$          \\ \midrule
L-MS-TCRNet                                           & $80.4 \pm 7.7$                        & $83.7 \pm 6.3$          & $87.7 \pm 8.2$          & $91.1 \pm 6.7$          & $88.2 \pm 9.2$          & $75.1 \pm 16.8$          \\
G-MS-TCRNet                                            & $\mathbf{80.9 \pm 8.0}$               & $\mathbf{84.1 \pm 6.7}$ & $\mathbf{89.0 \pm 7.7}$ & $\mathbf{91.7 \pm 6.3}$ & $\mathbf{88.7 \pm 9.4}$ & $\mathbf{75.5 \pm 17.6}$ \\ \midrule
\textbf{JIGSAWS}                                        & \multicolumn{1}{c}{\textbf{F1-macro}} & \textbf{Acc}            & \textbf{Edit}           & \textbf{F1@10}          & \textbf{F1@25}          & \textbf{F1@50}           \\ \midrule
BiLSTM* \cite{dipietro2019segmenting}  & \multicolumn{1}{c}{-}                 & $84.7$                  & $88.1$                  & -                       & -                       & -                        \\
BiGRU* \cite{dipietro2019segmenting}   & \multicolumn{1}{c}{-}                 & $84.8$                  & $88.5$                  & -                       & -                       & -                        \\
TCN + RL*  \cite{liu2018deep}          & \multicolumn{1}{c}{-}                 & $82.1$                  & $87.9$                  & $91.1$                  & $89.5$                  & $82.3$                   \\
APc \cite{van2020multi}                & \multicolumn{1}{c}{-}                 & $85.5$                  & $85.3$                  & -                       & -                       & -                        \\ \midrule
\textbf{\begin{tabular}[c]{@{}l@{}}G-MS-TCRNet+\\ +WFR+HI\end{tabular}}                                      & \multicolumn{1}{c}{$\mathbf{81.8 \pm 11.2}$}   & $\mathbf{86.4 \pm 7.3}$          & $\mathbf{90.5 \pm 11.4}$         & $\mathbf{94.1 \pm 7.5}$          & $\mathbf{93.6 \pm 8.1}$          & $\mathbf{87.0 \pm 13.1}$          \\ \bottomrule
\end{tabular}

}
\end{table}

In Supplementary Materials \ref{sub_sec:multi-modal}, we extend beyond kinematics-only models and compare our algorithm's performance with models based on video or multi-modal (kinematics + video) data evaluated on the JIGSAWS dataset.
Overall, our model outperforms all previous video-based algorithms and remains competitive with multi-modal networks.

\section{Conclusions}

 This work has a dual objective related to action segmentation tasks using kinematic data. First, we introduced two multi-stage architectures that achieved state-of-the-art results on kinematic data benchmark datasets. Second, we proposed two new augmentations for kinematic data, leveraging its strong geometric structure to enhance the performance and robustness of the algorithms.

 We evaluated our algorithms on three datasets: two open surgery simulations and one robotic surgery dataset. Our algorithms performed well regardless of whether the data were obtained from the da Vinci robot or electromagnetic sensors on the surgeon's hands. Despite differences in data acquisition methods, the preprocessing process is similar for both RAMIS and open surgery data.
 
Both architectures use a TCN-based prediction generator with intra-stage regularization for better frame-wise performance. Low-layer prediction heads focus on small areas, minimizing long-history impact. RNN-based refinement stages, with downsampling and upsampling, reduce over-segmentation errors. Downsampling reduces noise and optimizes frequencies for RNNs, while upsampling restores the original sequence dimensions.

Inspired by computer vision techniques, we developed two new augmentations for kinematic data: World-Frame Rotation and Hand Inversion. World-Frame Rotation, which randomly rotates the 3D coordinate frame, improved generalization, increasing mean performance and reducing standard deviation, especially in the complex BRS dataset. Hand Inversion addresses the under-representation of left-handed surgeons, significantly improving performance on their data without affecting right-handed surgeons. This led to higher mean performance and lower standard deviation.

Our method shows high performance in both robotic and open surgery. L-MS-TCRNet achieved the best results on the diverse BRS dataset, featuring 52 surgeons of varying handedness, with both augmentations contributing to its success. Additionally, G-MS-TCRNet outperformed the current state-of-the-art on the JIGSAWS and VTS benchmarks. On JIGSAWS, it also surpassed video-based algorithms and competed with multi-modal data algorithms.

Action segmentation is highly relevant across various domains where automated workflow analysis is essential, including skill assessment, process scheduling in industries, and error detection in work. Kinematic sensors can be utilized in these areas, similar to their use in surgical domains. Therefore, our method, showcased on datasets from open and robotic surgery, is not limited to the surgical domain and can be implemented in these additional domains as well.

However, our method has several possible limitations. First, the input data fed into our networks contains linear and angular velocities derived from the location and orientation of six sensors, all relative to the same world frame generated by the transmitter of the electromagnetic tracking system. An alternative to the electromagnetic tracking system is IMU sensors, which allow users to move freely while recording and analyzing data. Our architecture can be applied to IMU sensor data without any modifications. However, as IMUs measure linear acceleration and angular velocities relative to the body frame of each sensor, further preprocessing of the IMU data may be required for optimal performance.
Second, our augmentations assume the pose of all sensors relative to a common frame. This is applicable when using electromagnetic tracking systems (e.g. NDI's  Aurora and trackSTAR systems or Polhemus's VIPER, G4 PATRIOT systems). It is also relevant when using some IMU motion capture systems (e.g. Xsens's Awinda, BSN's Apex or NOITOM's Perception Neuron). However, it cannot be used as is when using the raw data from IMUs. 

We aim to address these limitations directly in future work. The electromagnetic tracking system provides position and orientation data, allowing us to simulate IMU data by calculating linear accelerations and angular velocities in the body frame. This enables performance comparison between IMU-like data and our current results on shared data.  This approach extends our work to IMU data, making it relevant to many new domains.

\section* {Acknowledgements}

The work was supported by the American College of Surgeons (Stanford Sponsored Project \# 162769) and National Institutes of Health (1R01DK12344501A1).

\section* {Declaration}
During the preparation of this work the authors used ChatGPT to rephrase sentences. After using this tool, the authors reviewed and edited the content as needed and take full responsibility for the content of the publication.

\bibliographystyle{elsarticle-num} 

\bibliography{ref.bib}

\newpage

\section {Supplementary Materials}

\subsection{BRS dataset analysis}
\label{sub_sec:BRS_dataset_analysis}

\begin{figure}[ht]
 \centering
 \includegraphics[width=1\textwidth]{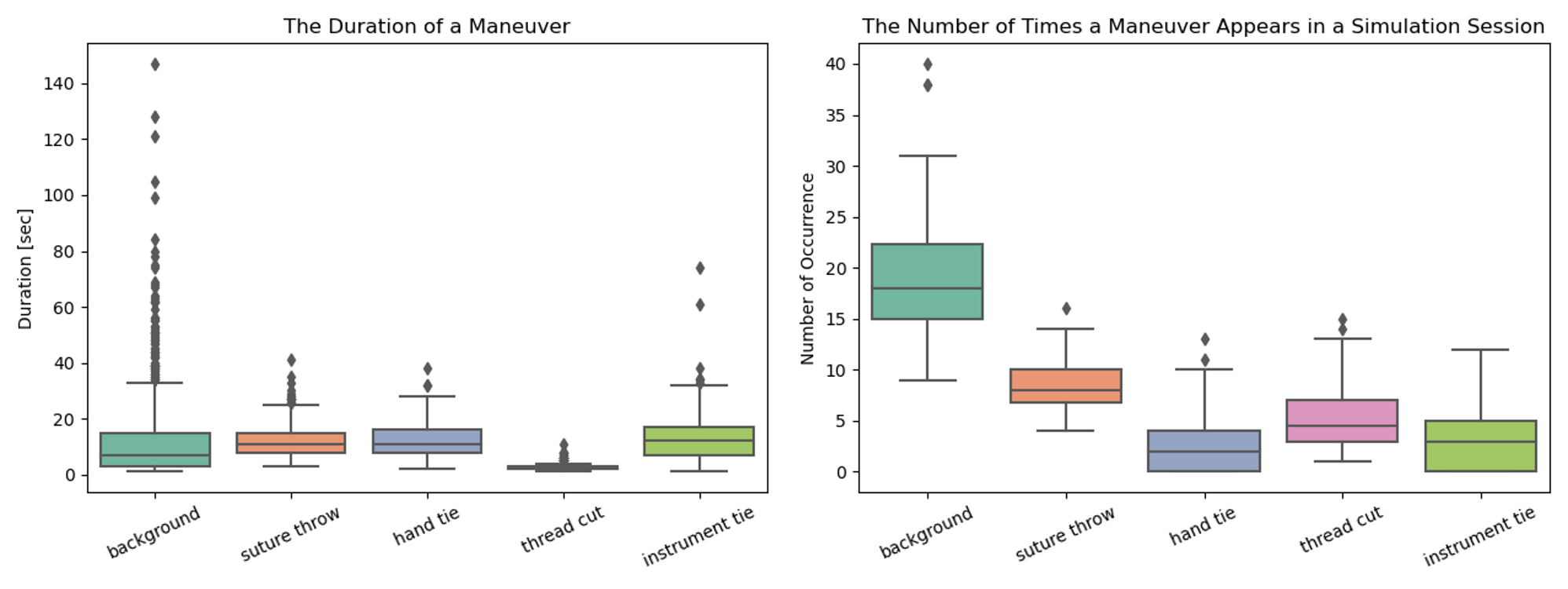}
 \caption{An illustration of the distributions of maneuvers in the BRS dataset in a box-plot with whiskers, drawn within the 1.5 IQR value. The left figure represents the distribution of the duration of a single instance of maneuver, and the right figure represents the distribution of the number of instances during a procedure of each type of maneuver.}
    \label{fig:BR_dist}

 \end{figure}

\Fig \ref{fig:BR_dist}
 presents a detailed dataset analysis, including the frequency and duration of each maneuver.

\subsection{VTS and BRS Data Acquisition and Prepossessing}
\label{subsec:acquisition_full}
Motion data was captured using electromagnetic motion sensors (NDI, trackSTAR Model 180). Three sensors were taped to the index, thumb, and wrist of each participant’s hands under the surgical gloves (see \Fig \ref{fig:datasets} -A).
Video data was captured using two cameras, one closeup camera focusing on the simulation area and one overview camera that included the surrounding area. The sensors and both cameras were captured and synchronized using MotionMonitor software (Innovative Sports Training, Inc.).
The labeling process relied on video data. Each activity, including gestures and maneuvers, was defined by its respective start and end times.
The raw data contains for each electromagnetic sensor three coordinates indicating the location of the sensor in space, and three Euler angles indicating the sensor's orientation for a total of $36$ kinematic variables for each timestamp. The original measurement rate was $179.695Hz$ in VTS and $100Hz$ in BRS datasets. 
The data from both datasets have been downsampled to $30Hz$ after filtering with an FIR low-pass filter. The filter was designed with the Parks-MacClellan algorithm \cite{mcclellan2005personal}. The pass-band cutoff frequency was $10Hz$ with a ripple of 3.91dB, and the stopband cutoff frequency was $15Hz$ with a ripple of -33.511dB.

\subsection{BiRNN formulation}
\label{sub_sec: BiRNN formulation}
\paragraph{Basic RNN Units}
Let us now formally define $BiLSTM(\cdot)$ and $BiGRU(\cdot)$, which are practical realizations of $RNN(\cdot)$ in equation \ref{eq:RNN}.
First, let's define LSTM and GRU units.
One of the most widely adopted RNN architectures is the long short-term memory (LSTM), which addresses the vanishing gradient problem \cite{hochreiter1997long}. The LSTM unit can be formalized as follows:
    \begin{align}
    i_{t}&=\sigma(W_{ii}x_{t}+W_{hi}h_{t-1}+b_{i}) \\
    f_{t}&=\sigma(W_{if}x_{t}+W_{hf}h_{t-1}+b_{f}) \\
    g_{t}&=tanh(W_{ig}x_{t}+W_{hg}h_{t-1}+b_{g}) \\
    o_{t}&=\sigma(W_{io}x_{t}+W_{ho}h_{t-1}+b_{o}) \\
    c_{t}&=f_{t}\odot c_{t-1}+i_{t}\odot g_{t} \\
    h_{t}&=o_{t}\odot tanh(c_{t})
    \end{align}
    Where $\odot$ is the Hadamard product, $\sigma$ and $tanh$ are the sigmoid and hyperbolic tangent activation functions.
    A hidden state at time t is represented by $h_t \in \Reals^{Q}$, a cell state is represented by $c_t \in \Reals^{Q}$, an input is represented by $x_t \in R^{M}$ where $M$ is the input vector dimension, and a hidden state at time t-1 is represented by $h_{t-1}$. Input, forget, cell, and output gates are represented by $i_t \in \Reals^{Q}$, $f_t  \in \Reals^{Q}$ , $g_t  \in \Reals^{Q}$, and $o_t$ respectively;  $W_{ii} \in \Reals^{Q \times M}$, $W_{if} \in \Reals^{Q \times M}$, $W_{ig} \in \Reals^{Q \times M}$,$W_{io} \in \Reals^{Q \times M}$, $W_{hi} \in \Reals^{Q \times Q}$, $W_{hf} \in \Reals^{Q \times Q}$, $W_{hg} \in \Reals^{Q \times Q}$, and $W_{ho} \in \Reals^{Q \times Q}$ are matrices of weights; and $b_{i} \in \Reals^{Q}$, $b_{f} \in \Reals^{Q}$ , $b_{g} \in \Reals^{Q}$, and $b_{o} \in \Reals^{Q}$ are biases.\\
    
    A popular alternative to LSTM is gated recurrent units (GRUs) \cite{GRU2014}. GRUs are capable of alleviating the vanishing gradient problem in a very similar manner to LSTMs but are simpler and require fewer parameters and operations. 
     
    \begin{align}
        r_{t}&=\sigma(W_{ir}x_{t}+W_{hr}h_{t-1}+b_{r}) \\
        z_{t}&=\sigma(W_{iz}x_{t}+W_{hz}h_{t-1}+b_{z}) \\
        n_{t}&=tanh(W_{in}x_{t}+b_{in}+r_{t} \odot (W_{hn}h_{t-1}+b_{hn})) \\ h_{t}&=(1-z_{t}) \odot n_{t}+z_{t} \odot h_{t-1}
    \end{align}
    Where $r_t \in \Reals^{Q}$, $z_t \in \Reals^{Q}$, $n_t \in \Reals^{Q}$ are the reset, update, and new gates, respectively; $W_{ir} \in \Reals^{Q \times M}$, $W_{iz} \in \Reals^{Q \times M}$, $W_{in} \in \Reals^{Q \times M}$,
    $W_{hr} \in \Reals^{Q \times Q}$,
    $W_{hz} \in \Reals^{Q \times Q}$, and
    $W_{hn} \in \Reals^{Q \times Q}$ are weights matrices;
    and $b_{r} \in \Reals^{Q}$, $b_{z} \in \Reals^{Q}$, $b_{in} \in \Reals^{Q}$, and $b_{hn} \in \Reals^{Q}$ are bias vectors.

\paragraph{RNNs with Extended Structures}
    We have implemented each architecture in a bidirectional manner with multiple RNN layers.
    In order to obtain a bidirectional structure, one RNN has to run in the forward direction, while the other one has to run in the reverse direction. An output of the bidirectional structure is the concatenation of the hidden states of both RNNs that correspond to the same time.
    In the single-layer case, the hidden state $h_{t}$ of time $t$ serves as an output of this unit, whereas in the multi-layer case, the input of $x^{(\ell)}_t$ of the $\ell^{th}$ layer is the hidden state $h^{(\ell-1)}_t$ of the previous layer.

\subsection{Hyperparameter Tuning}
\label{sub_sec:Hyperparameter_Tuning_full}
    The hyperparameter optimization was executed only on the VTS dataset on both our L-MS-TCRNet and G-MS-TCRNet networks using the Optuna environment \cite{optuna_2019} with the TPESampler sampler for 300 trials.
    The following hyperparameters were tuned and associated with the prediction generator (PG): TCN dropout probability in the range of $[0.5,0.7]$, number of layers in the PG out of $\{7,9,11,13\}$, number of feature maps in the PG out of $\{32,64,128,256\}$.
    The following hyperparameters were tuned and associated with the refinement module: 
    number of refinement stages out of $\{1,2,3\}$, RNN dropout probability in the range of $[0.5,0.7]$,
    RNN hidden dimension size out of $\{128,256,512\}$, number of RNN layers out of $\{1,2\}$.
    The following hyperparameters were tuned and associated with the entire network:
    learning rate in the range of $[10^{-4},5 \cdot 10^{-3}]$, loss hyper parameter $\lambda$ in the range of $[0.15,1]$, batch size in the range of $[1,5]$,
    primary sampling factor $r$ in the range of $[1,8]$, and secondary sampling factor in the range of $[1,\lfloor 8 / r \rfloor]$, where the sampling rates are factors that define how many samples are required to skip. The primary sampling factor is the downsample that is applied to the input of the network, and the secondary sampling factor refers to the downsampling performed in the refinement stages on the previous stage output. The number of epochs was 40; Adam optimizers were used to train all networks with $\beta1=0.9$ and $\beta2=0.999$.
    Eight Nvidia A100 GPUs were used for training and evaluation on a DGX Cluster.
    The optimal hyperparameters for each of our networks are described in \Table \ref{Table:Params}. The optimized L-MS-TCRNet network achieved a mean F1-Macro of $80.9$ on the validation set with $6.3 \times 10^{6}$ parameters and the obtained G-MS-TCRNet attained a mean F1-Macro of $81.2$ with $8.4 \times 10^{6}$ parameters.
    On all datasets, we used the optimized networks as described in this section without any additional hyperparameter tuning.\\\\

\begin{table}[ht]
\centering
\caption{Selected hyperparameters based on the VTS validation set, used for all other evaluations as-is.}
\label{Table:Params}
\resizebox{\columnwidth}{!}{%
\begin{tabular}{@{}lcc@{}}
\toprule
 & \multicolumn{1}{l}{\textbf{L-MS-TCRNet}} & \multicolumn{1}{l}{\textbf{G-MS-TCRNet}} \\ \midrule
Number of layers (PG) & 11 & 13 \\
Number of feature maps (PG) & 256 & 256 \\
Dropout (PG) & 0.546 & 0.645 \\
Number of refinement stages (Refinement) & 1 & 1 \\
Number of layers (Refinement) & 2 & 2 \\
Dropout (Refinement) & 0.619 & 0.5747 \\
Hidden dimension size (Refinement) & 128 & 256 \\
Learning rate & 0.001035 & 0.001779 \\
Primary sampling rate & 2 & 1 \\
Secondary sampling rate & 3 & 6 \\
$\lambda$ & 0.933 & 0.638 \\
Number of parameters & $6.3 \times 10^{6}$ & $8.4 \times 10^{6}$ \\ \bottomrule
\end{tabular}%
}
\end{table}

\subsection{comparison with video or multi-modal networks}
\label{sub_sec:multi-modal}

In \Table \ref{Table:JIGSAWS-multimodal}, we extend beyond kinematics-only models and compare our algorithm's performance with models based on video or multi-modal (kinematics + video) data evaluated on the JIGSAWS dataset.
Our G-MS-TCRNet with WFR+HI augmentation achieves second place in Edit distance and F1@10 scores, showing competitiveness with recent state-of-the-art results in a multimodal architecture combining video and kinematic data published in 2022 \cite{van2022gesture}. In terms of accuracy, we secure third place, with a 1.5-point difference from the state-of-the-art.
Additionally, we achieve state-of-the-art results for F1@25 and F1@50, although not all recent publications report these metrics. Overall, our model outperforms all previous video-based algorithms and remains competitive with multi-modal networks.

\begin{table}[ht]
\centering
\caption{Comparison of our algorithms, which were trained on kinematic data to state-of-the-art results on the JIGSAWS dataset, to models trained on video and multi-modal data. Models marked with an * are based on the original JIGSAWS labels. Models marked with $\dag$ were implemented by \cite{wang2020towards}. During the training of our models, we used WFR+HI augmentation. The WFR augmentation was applied with a probability of one, with $\theta_{max} =7 \degree$. HI was applied with a probability of 0.5. The best results are in bold, the second place is marked by underlining.}
\label{Table:JIGSAWS-multimodal}
\scalebox{0.7}{

\begin{tabular}{@{}lllccccc@{}}
\toprule
\multirow{2}{*}{\textbf{}}                                                                                    & \multicolumn{2}{c}{\textbf{Modality}}                 & \multirow{2}{*}{\textbf{Acc}} & \multirow{2}{*}{\textbf{Edit}} & \multirow{2}{*}{\textbf{F1@10}} & \multirow{2}{*}{\textbf{F1@25}} & \multirow{2}{*}{\textbf{F1@50}} \\ \cmidrule(lr){2-3}
                                                                                                              & \multicolumn{1}{c}{Kin}   & \multicolumn{1}{c}{Vid}   &                               &                                &                                 &                                 &                                 \\ \midrule
\textbf{C3D-MTL-VF* \cite{wang2020towards}}                                                  &                           & \checkmark & 82.1                          & 86.6                           & 90.6                            & 89.1                            & 80.3                            \\
\textbf{MS-TCN* \cite{farha2019ms} (implemented by \cite{wang2020towards})} &                           & \checkmark & 78.9                          & 85.8                           & 88.5                            & 86.6                            & 75.8                            \\
\textbf{TCN + RL*  \cite{liu2018deep}}                                                       &                           & \checkmark & 81.4                          & 88                             & 92                              & 90.5                            & 82.2                            \\
\textbf{TDRN* \cite{lei2018temporal}}                                                        &                           & \checkmark & 84.6                          & 90.2                           & 92.9                            & -                               & -                               \\
\textbf{RL+Tree* \cite{gao2020automatic}}                                                    &                           & \checkmark & 81.7                          & 88.5                           & 92.7                            & 91.0                            & 83.2                            \\
\textbf{MRG-Net* \cite{long2021relational}}                                                  & \checkmark & \checkmark & \textbf{87.9}                          & 89.3                           & -                               & -                               & -                               \\
\textbf{Fusion-KV* \cite{qin2020temporal}}                                                   & \checkmark & \checkmark & 86.3                          & 87.2                           & -                               & -                               & -                               \\
\textbf{MA-TCN \cite{van2022gesture}}                                                        & \checkmark & \checkmark & \ul{86.8}                 & \textbf{91.4}                  & \textbf{94.3}                   & -                               & -                               \\ \midrule
\textbf{\begin{tabular}[c]{@{}l@{}}G-MS-TCRNet+\\ +WFR+HI\end{tabular}}                                      & \checkmark &                           & 86.4                  & \ul{90.5}                     & \ul{94.1}                      & \textbf{93.6}                   & \textbf{87.0}                   \\ \bottomrule
\end{tabular}

}
\end{table}

\end{document}